\documentclass[journal]{IEEEtran}
\IEEEoverridecommandlockouts
\usepackage{cite}
\usepackage{hyperref}
\usepackage{algorithm,algorithmic}
\usepackage{setspace}
\usepackage{stfloats}
\usepackage{amsmath,amssymb,amsfonts}
\usepackage{algorithmic}
\usepackage{amsmath}
\usepackage{mathtools}
\usepackage{booktabs}
\usepackage{graphicx}
\usepackage{multirow}
\usepackage[capitalise]{cleveref}
\usepackage{fontawesome}
\usepackage{graphicx}
\usepackage{subfigure}
\usepackage{xcolor}
 
\usepackage{graphicx}
\usepackage{subfigure}
\usepackage{amsmath}
\usepackage{xcolor}
\usepackage{color, soul}
\def\BibTeX{{\rm B\kern-.05em{\sc i\kern-.025em b}\kern-.08em
    T\kern-.1667em\lower.7ex\hbox{E}\kern-.125emX}}
\begin{document}

\title{
% A Reliable Optimal Power Flow Solver by Neural Networks Integrating Physics Models and worth-learning Data Generation\\
% A Reliable Optimal Power Flow Solver by Worth-learning Data Generation Based on Physical-model-integrated Neural Network
Optimal Power Flow Based on Physical-Model-Integrated Neural Network with Worth-Learning Data Generation
}

\author{Zuntao Hu,~\IEEEmembership{Graduate Student Member,~IEEE,}
and Hongcai Zhang,~\IEEEmembership{Member,~IEEE}
}

\maketitle

\begin{abstract}
Fast and reliable solvers for optimal power flow (OPF) problems are attracting surging research interest.
As surrogates of physical-model-based OPF solvers, neural network (NN) solvers can accelerate the solving process. However, they may be unreliable for ``unseen" inputs when the training dataset is unrepresentative. Enhancing the representativeness of the training dataset for NN solvers is indispensable but is not well studied in the literature. To tackle this challenge, we propose an OPF solver based on a physical-model-integrated NN with worth-learning data generation. The designed NN is a combination of a conventional multi-layer perceptron (MLP) and an OPF-model module, which outputs not only the optimal decision variables of the OPF problem but also the constraints violation degree. Based on this NN, the worth-learning data generation method can identify feasible samples that are not well generalized by the NN. By iteratively applying this method and including the newly identified worth-learning samples in the training set, the representativeness of the training set can be significantly enhanced.
Therefore, the solution reliability of the NN solver can be remarkably improved. 
Experimental results show that the proposed method leads to an over 50\% reduction of constraint violations and optimality loss compared to conventional NN solvers.
\end{abstract}


\section{Introduction}
\IEEEPARstart{O}{ptimal} power flow (OPF) is a fundamental but challenging problem for power systems \cite{taylor2015convex}. A typical OPF problem usually involves determining the optimal power dispatch with an objective, e.g., minimizing total generation costs or power loss, while satisfying nonlinear power flow equations and other physical or engineering constraints \cite{OPF-book}. Due to the nonlinear interrelation of nodal power injections and voltages, OPF is non-convex, NP-hard, and cannot be solved efficiently \cite{OPF_NP}. With the increasing integration of renewable generation and flexible demands, uncertainty and volatility have been rising on both the demand and supply sides of modern power systems \cite{tang2017real}, which requires OPF to be solved more frequently. Thus, fast and reliable OPF solvers have become indispensable to ensure effective operations of modern power systems and have attracted surging interest in academia. 

There is a dilemma between the solving efficiency and solution reliability of OPF. Conventionally, OPF is  solved by iterative algorithms, such as interior point algorithms, based on explicit physical models \cite{Matpower}. However, these methods may converge to locally optimal solutions. Recently, some researchers have made great progress in designing conic relaxation models for OPF, which are convex and can be efficiently solved \cite{jabr2006radial,bai2008semidefinite,lavaei2011zero}. Nevertheless, the exactness of these relaxations may not hold in practical scenarios, and they may obtain infeasible solutions \cite{Low2014Convex}. In addition, the scalability of the conic relaxation of alternating current optimal power flow (AC-OPF) may still be a challenge, particularly in online, combinatorial, and stochastic settings \cite{sen-inf}. 

To overcome the limitation of the aforementioned physical-model-based solvers, some researchers propose surrogate OPF solvers based on neural networks (NNs) \cite{Deep0opf, OPFlarg, Deep1OPF}. These solvers use NNs to approximate the functional mapping from the operational parameters (e.g., profiles of renewable generation and power demands) to the decision variables (e.g., power dispatch) of OPF. Compared to iterative algorithms, they can introduce significant speedup because an NN is only composed of simple fundamental functions in sequence \cite{OPFlarg, Deep1OPF}. However, one of the critical problems of NN solvers is that they may be unreliable if not properly trained, especially for ``unseen" inputs in feasible regions due to NNs' mystery generalization mechanism \cite{Wang2022Generalizing}.

The generalization of NNs is mainly influenced by their structures, loss functions, and training data. Most published papers propose to enhance the generalization of NN OPF solvers by adjusting the structures and loss functions. Various advanced NN structures rather than conventional fully connected networks are employed to imitate AC-OPF. For example, Owerko \textit{et al.} \cite{owerkoOptimalPowerFlow2020} use graph NNs to approximate a given optimal solution. Su \textit{et al.} \cite{su2021optimized} employ a deep belief network to fit the generator's power in OPF. Zhang \textit{et al.} \cite{zhang2021dcopfGuarantee} construct a convex NN solving DC-OPF to guarantee the generalization of NNs. Jeyaraj \textit{et al.} \cite{JEYARAJ2022107730Optimum} employ a Bayesian regularized deep NN to solve the OPF in DC microgrids.
Some researchers design elaborate loss functions that penalize the constraints violation, combine Karush-Kuhn-Tucker conditions, or include derivatives of decision variables to operational parameters. For example, Pan \textit{et al.} \cite{Deep0opf} introduce a penalty term related to the inequality constraints into the loss function. This approach can speed up the computation by up to two orders of magnitude compared to the Gurobi solver, but 18.3\% of its solutions are infeasible. Ferdinando \textit{et al.} \cite{OPFlarg} include a Lagrange item in the loss function of NNs. Their method's prediction errors are as low as 0.2\%, and its solving speed is faster than DC-OPF by at least two orders of magnitude. Manish \textit{et al.} \cite{sen-inf} include sensitivity information in the training of NN so that only using about 10\% to 25\% of training data can attain the same approximation accuracy as methods without sensitivity information. Nellikkath \textit{et al.} \cite{phy-inf} apply physics-informed NNs to OPF problems, and their results have higher accuracy than conventional NNs.

The above-mentioned studies have made significant progress in designing elaborate network structures and loss functions. However, little attention has been paid to the training set generation problem. Specifically, they all adopt conventional probability sampling methods to produce datasets for training and testing, such as simple random sampling \cite{Huang2022Dopfv, Deep0opf, Deep1OPF,sen-inf, OPFlarg,zhang2021dcopfGuarantee, owerkoOptimalPowerFlow2020}, Monte Carlo simulation \cite{JEYARAJ2022107730Optimum}, or Latin hypercube sampling \cite{su2021optimized, phy-inf}.
These probability sampling methods cannot provide a theoretical guarantee that a generated training set can represent the input space of the OPF problem properly. As a result, probability sampling methods may generate insufficient and unrepresentative training sets, so the trained NN solvers may provide unreliable solutions. 
% A training set's capacity to represent the input space significantly impacts the NNs' generalization. Therefore, these probability sampling methods, which may lead to a lack of samples in some input-place sub-regions, hinder NNs from generalizing to the sub-regions.
 
It is important to create a sufficiently representative dataset for training an NN OPF solver. A training set's representativeness depends on its size and distribution in its feasible region \cite{naMa}. Taking a medium-scale OPF problem as an example,  millions of data samples may still be sparse given the high dimension of the NN's inputs (e.g., operational parameters of the OPF problem: renewable generation and power demands at all buses); in addition, because the OPF problem is non-convex, the feasible region of the NN's inputs is a complicated irregular space. Thus, generating a representative training set to cover all the feasible regions of the inputs with an acceptable size is quite challenging. Without a representative training set, it is difficult to guarantee that the NN OPF solver's outputs are reliable, especially given ``unseen" inputs in the inference process, as discussed in \cite{poVuner,VerGuarantee}. 

% Moreover, the DNN solver can always generate a solution, and this feature will lead to an unexpected result if the input is out of its feasible region.

To address the above challenge, this study proposes a physical-model-integrated deep NN method with worth-learning data generation to solve AC-OPF problems. To the best of our knowledge, this is the first study that has addressed the representativeness problem of the training dataset for NN OPF solvers.
% This paper proposes an approach to obtain a reliable DNN solver by a novel training-data generation method, where each sample is produced if it carries more information than existing ones, meaning the data is worth learning. So the training set's information would gradually increase to the required level of NNs generalizing to the feasible region. \textcolor{red}{Moreover, specific NNs are designed not only for the worth-learning data generation method but also for more minor violation degrees with the same training loss.}
The major contributions of this study are twofold:
\begin{enumerate}
    \item A novel physical-model-integrated NN is designed for solving the AC-OPF problem. This NN is constructed by a conventional MLP integrating an OPF-model module, which outputs not only the optimal decision variables of the OPF problem but also the violation degree of constraints. By penalizing the latter in the loss function during training, the NN can generate more reliable decision variables.
    \item Based on the designed NN, a novel generation method for worth-learning training data is proposed, which can identify samples in the input feasible region that are not well generalized by the previous NN. By iteratively applying this method during the training process, the trained NN  gradually generalizes to the whole feasible region. As a result, the generalization and reliability of the proposed NN solver can be significantly enhanced.
\end{enumerate}
Furthermore, comprehensive numerical experiments are conducted, which prove that the proposed method is effective in terms of both reliability and optimality for solving AC-OPF problems with high computational efficiency.
%but also indicate whether the result is feasible or not

The remainder of this article is organized as follows. Section \ref{Section 2} provides preliminary models and the motivations behind this study. Section \ref{Section 3} introduces the proposed method. 
% Verification of convergence and efficiency of the proposed method is in Section \ref{Section 4}. 
Section \ref{Section 5} details the experiments. Section \ref{Section 6} concludes this paper. 

\section{Analysis of Approximating OPF Problems by NN}
\label{Section 2}
\subsection{AC-OPF problem}
The AC-OPF problem aims to determine the optimal power dispatch (usually for generators) given specific operating conditions of a power system, e.g., power loads and renewable generation. A typical AC-OPF model can be formulated as %\cite{Matpower}
\begin{subequations}
\label{AC-OPF model}
    \begin{align}
	\underset{\boldsymbol{V},\, \boldsymbol{S}^\text{G}}{\min}\,\,&\mathrm{C}\,\left(\boldsymbol{S}^\text{G} \right) \,\quad \,\, \label{obj}\\
	\,\,\,\text{s.t.:}~~ &\left[ \boldsymbol{V} \right] \mathbf{Y}_\text{bus}^{*}\boldsymbol{V}^*=\boldsymbol{S}^\text{G}-\boldsymbol{S}^\text{L},\quad \,\,\quad  \label{c1}\\
	&\underline{\mathbf{S}}^\text{G}\le \boldsymbol{S}^\text{G}\le \overline{\mathbf{S}}^\text{G},\quad \,\,\quad \label{c2}\\
	&\underline{\mathbf{V}}\le \boldsymbol{V}\le \overline{\mathbf{V}},\quad \label{c3}\\
	&|\mathbf{Y}_\text{b} \boldsymbol{V}|\le \mathbf{\bar{I}},
	\label{c4} 
    \end{align}
\end{subequations}
where \cref{obj} is the objective, e.g., minimizing total generation costs, and \cref{c1,c2,c3,c4} denote constraints. Symbols $\boldsymbol{S}^\text{G}$ and $\boldsymbol{S}^\text{L}$ are $n \times1$ vectors representing complex bus injections from generators and loads, respectively, where $n$ is the number of buses. Symbol $\boldsymbol{V}$ is an $n \times1$ vector denoting node voltages. Symbol $[.]$ denotes an operator that transforms a vector into a diagonal matrix with the vector elements on the diagonal. Symbol $\textbf{Y}_\text{bus}$ is a complex $n \times n$ bus admittance matrix written as $\textbf{Y}$ at other sections for convenience. Symbol $\textbf{Y}_\text{b}$ is a complex $n_\text{b} \times n$ branch admittance matrix, and $n_\text{b}$ is the number of branches. The upper and lower bounds of any variable $\boldsymbol{x}$ are represented by $\bar{\boldsymbol{x}}$ and $\underline{\boldsymbol{x}}$, respectively. Vector $\mathbf{\bar{I}}$ denotes the current flow limit of branches.

\subsection{{AC-OPF mapping from loads to optimal dispatch}}
\label{mappingF}
An NN model describes an input-output mapping. Specifically, for an NN model solving the AC-OPF problem shown in \cref{AC-OPF model}, the input is the power demand $\boldsymbol{S}^\text{L}$, and the output is the optimal generation $\boldsymbol{S}^\text{G\,*}$. Hence, an NN OPF solver describes the mapping $\boldsymbol{S}^\text{G\,*} = f^\text{OPF}(\boldsymbol{S}^\text{L})$. A well-trained NN should be able to accurately approximate this mapping. 
\begin{figure}\vspace{-5mm}
    \centering
    \includegraphics[width=2.4in]{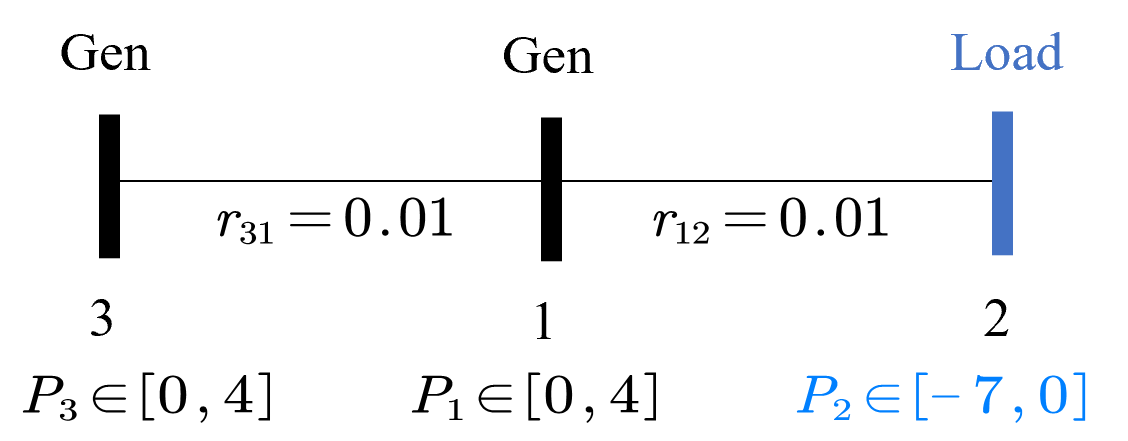}
    \caption{The 3-bus system.}\vspace{-5mm}
    \label{3-bus}
\end{figure}

We provide a basic example of a 3-bus system, as shown in \cref{3-bus}, to illustrate how NN works for OPF problems and explain the corresponding challenge for generalization. For simplicity, we assume there is no reactive power in the system and set $r_{31} =r_{12}= 0.01$
; $\underline{P}_{i}=0$ , $\overline{P}_{i}=4$, for $i\in \{1,3\}$, $P_{2}\in[-7, 0]$; $\underline{V}_{i}=0.95$ and $\overline{V}_{i}=1.05$, for $i\in \{1,2,3\}$.  Then, the OPF model \cref{AC-OPF  model} is reduced to the following quadratic programming:
\begin{subequations}
\begin{align}
\underset{\textbf{\textit{V}},\, \textbf{\textit{P}}^\text{G}}{\min}\,\, &\ P_{1}+1.5\times P_{3}  \label{obj-3bus}
\\
\,\,\,  \text{s.t.:} \,\,  &P_{1}=V_1\left( V_1-V_2 \right)   / 0.01 +V_1\left( V_1-V_3 \right) / 0.01,
\\
\,\,      &P_{2 }=V_2\left( V_2-V_1 \right)  / 0.01,
\\
\,\,      &P_{3 }=V_3\left( V_3-V_1 \right)  / 0.01,
\\
\,\,       &0.95\le V_3\le 1.05,  0.95\le V_2\le1.05,
\\
\,\,       &0\le P_{1 } \le 4, 0 \le P_{3 }\le 4, V_1=1, 
\end{align}
\label{3_bus}
\end{subequations}
where $\boldsymbol{V}$ is $[V_1 \,\,\, V_2\,\,\,V_3]^{\top}$, and $\boldsymbol{P}^\text{G}$ is $[P_1\,\, P_3]^{\top}$.

Given that $P_2$ ranges from -7 to 0, the 3-bus OPF model can be solved analytically. The closed-form solution of $[P_1^*\,\, P_3^* ]$ = $f^\text{OPF}_{\text{3-bus}}$($P_2$), is formulated as follows:
\begin{equation} \label{condi1}
P_1^*=\left\{ \begin{matrix}
	50-50\sqrt{0.04P_2+1},&		\mathrm{c}1,\\
	4,&		\mathrm{c}2,\\
\end{matrix} \right. 
\end{equation}
\begin{equation} \label{condi2}
P_3^*=\left\{ \begin{matrix}
	0,&		\mathrm{c}1,\\
	\begin{array}{c}
	213\left( 1-0.34\sqrt{0.04P_2+1} \right) ^2\\
	+50\sqrt{0.04P_2+1}-146\\
\end{array},&		\mathrm{c}2,\\
\end{matrix} \right. 
\end{equation}
where \text{c1} denotes condition 1: ~$-3.84\le P_{2}\le 0$, and \text{c2} denotes condition 2: $-7\le P_{2}< -3.84$.

\begin{figure}
    \centering
    \includegraphics[width=3.5in]{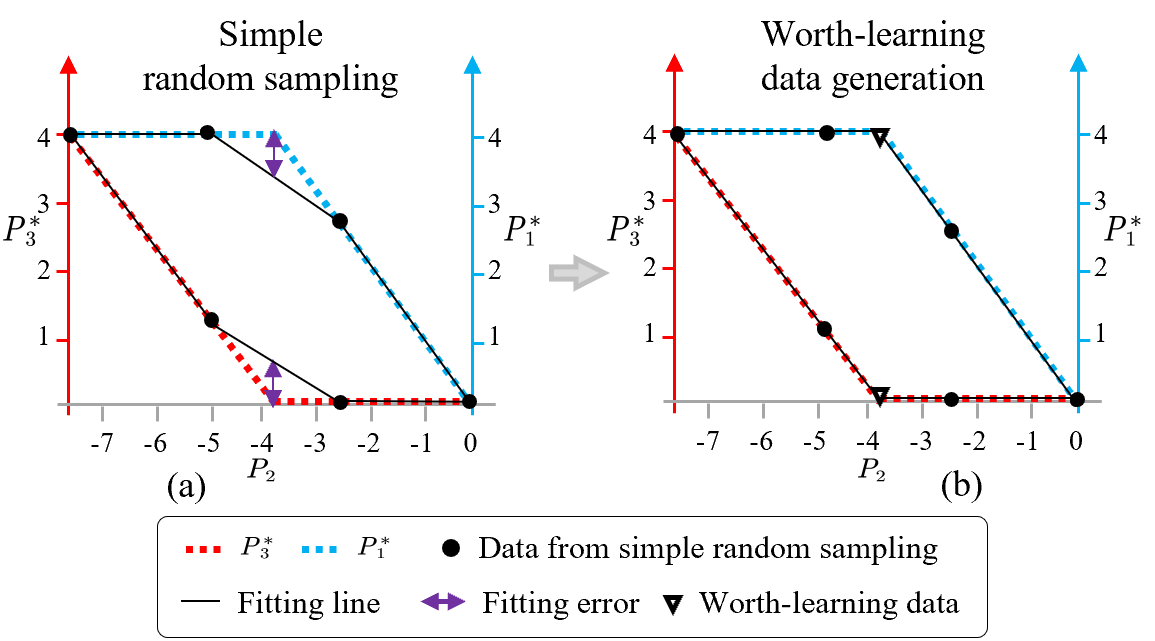}
    \vspace{-4mm}
    \caption{Examples of an NN fitting the OPF of the 3-bus system based on (a) simple random sampling, and (b) worth-learning data generation.}\vspace{-5mm}
    \label{MMP}
\end{figure}
To further analyze the mapping $f^\text{OPF}_{\text{3-bus}}$, we draw the $[P_1^*\,\, P_3^* ] \text{--}P_{2}$ curve according to \cref{condi1,condi2}, shown in \cref{MMP}. {Both the $P_1^* \text{--}P_{2}$ and $P_3^*  \text{--}P_{2}$ curves are piecewise nonlinear functions, in which two oblique lines are nonlinear because of the quadratic equality constraints.}
The reason why the two curves above are piecewise is that the active inequalities change the $[P_1^*\,\, P_3^* ] \text{--}P_{2}$ relationship.
From an optimization perspective, each active inequality will add a unique equality constraint to the relationship, so the pieces in  $f_\text{3-bus}^\text{OPF}$ are determined by the sets of active inequalities. In this example, the two pieces in each curve correspond to two sets of active inequalities: $P_{1 } \le 4 $ and $0 \le P_{3}$. Moreover, the two intersection points are the critical points where these inequalities are just satisfied as equalities.

For a general AC-OPF problem, its input is usually high-dimensional (commonly determined by the number of buses), and its feasible space is partitioned into some distinct regions by different sets of active inequality constraints. From an optimization perspective, a set of active constraints uniquely characterizes the relationship $\boldsymbol{S}^\text{G\,*} = f^\text{OPF}(\boldsymbol{S}^\text{L})$, and the number of pieces theoretically increases with the number of inequality constraints by exponential order \cite{MPP, MPP2, MPP3}. Therefore, there are massive regions, and each region corresponds to a unique mapping relation, i.e., a piece of mapping function $f^\text{OPF}$.  

\subsection{Challenges of fitting OPF mapping by NN}
As shown in \cref{MMP}(a), to fit the two-dimensional piecewise nonlinear curve of  $f^\text{OPF}_{\text{3-bus}}$, we first adopt four data samples by simple random sampling and then use an NN to learn the curve. Obviously, there are significant fitting errors between the fitting and the original lines. Because the training set lacks the samples near the intersections in the curve (where $p_2=-0.384$ in this case), the NN cannot accurately approximate the mapping in the neighboring region of the intersections. 

A training set representing the whole input space is a prerequisite for an NN approximating the curve properly. However, it is nontrivial to generate a representative training set by probability sampling. As shown in \cref{MMP}(a), the intersections of $f^{\text{OPF}}$ are key points for the representativeness, and the number of intersections  increases exponentially with that of the inequality constraints, as analyzed in \ref{mappingF}. When each sample is selected with a small possibility $\rho$, the generation of a dataset containing all the intersection points are in a low possibility event whose probability is equal to $\rho^m$, where $m$ is the number of intersections. In practice, the only way to collect sufficient data representing the input space by probability sampling is to expand the dataset as much as possible \cite{OPF-Learn}. This is impractical for large power networks. Therefore, the conventional probability sampling in the literature can hardly produce a representative dataset with a moderate size.

As shown in \cref{MMP}(b), if we are able to identify the two intersections, i.e., ($P_2=-0.384$, $P_1=4$) and ($P_2=-0.384$, $P_3=0$), and include them as new samples in the training dataset, the corresponding large fitting errors of the NN can be eliminated. These samples are termed as the \textit{worth-learning} data samples. The focus of this study is to propose a \textit{worth-learning data generation method} that can help identify worth-learning data samples and overcome the aforementioned disadvantage of conventional probability sampling (detailed in the following section).
% \textcolor{red}{ For the trained NN from an unrepresentative dataset (\cref{MMP}(a)), the data generation method is able to identify the worth-learning data, which leads to large fitting errors of the current NN. Then adding worth-learning data to the training set, the modified NN eliminates those fitting errors by learning on the new dataset (\cref{MMP}(b)).}
% With every data sample worth learning, a moderate-size dataset  representing the whole feasible region can be produced. As a result, the training performance of the NN AC-OPF solver can be significantly enhanced. 

\section{A Physical-Model-Integrated NN with Worth-Learning Data Generation}
\label{Section 3}

This section proposes a physical-model-integrated NN with a worth-learning data generation method to solve AC-OPF problems.
The proposed NN is a combination of a fully-connected network and a transformed OPF model. It outputs not only the optimal decision variables of the OPF problem but also the violation degree of constraints, which provides guidance for identifying worth-learning data. The worth-learning data generation method
creates representative training sets to enhance the generalization of the NN solver. 
\subsection{Framework of the proposed method}
\begin{figure}[tb]
    \centering
    \includegraphics[width=3.0in]{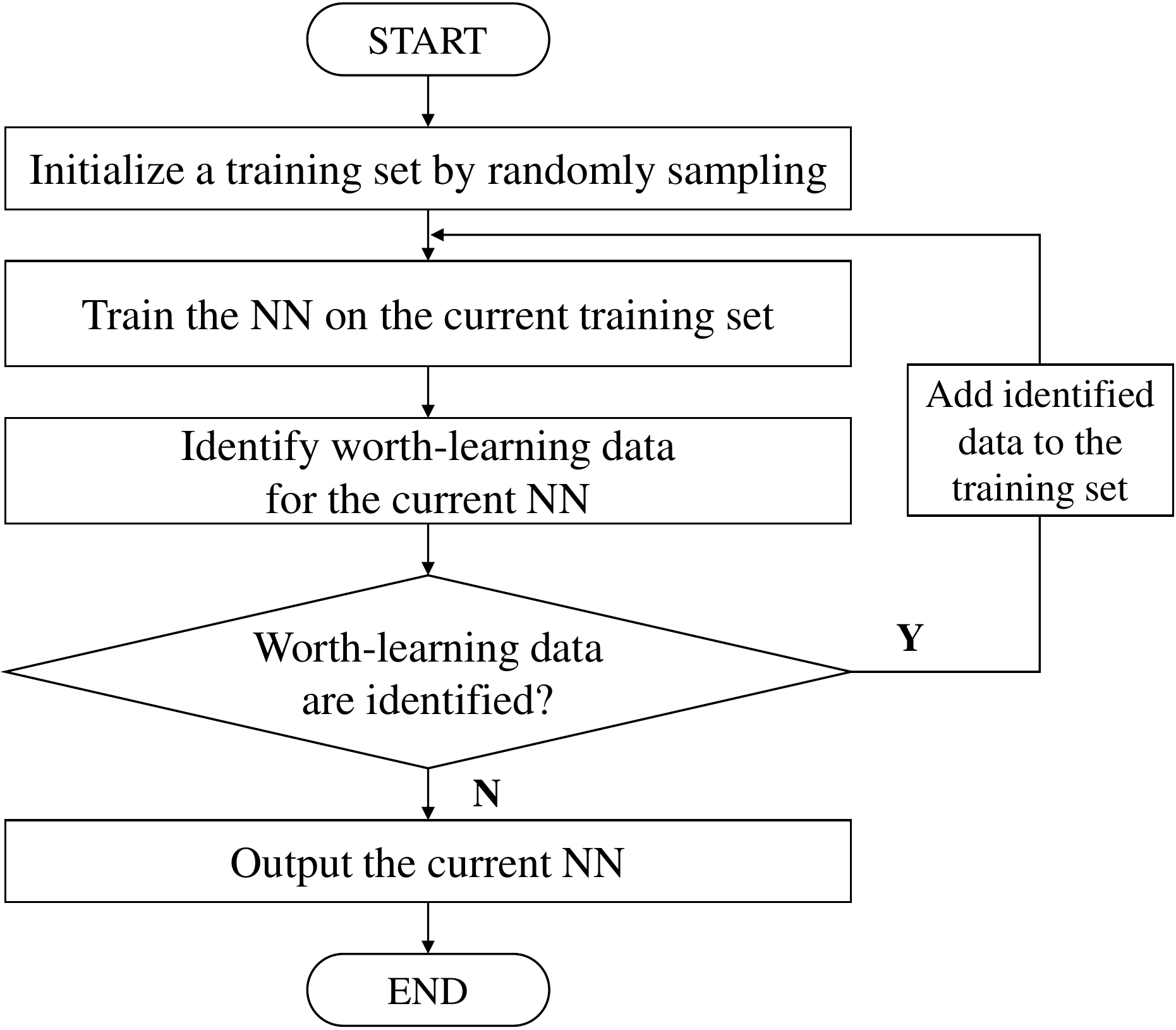}
    \caption{Framework of the proposed training process.}
    \label{trainflowchart} \vspace{-5mm}
\end{figure}

The proposed data generation method has an iterative process, as shown in \cref{trainflowchart}. First,  a training set is initialized by random sampling; second,  the physical-model-integrated NN is trained on the training set, where an elaborate loss function is utilized; third,  worth-learning data for the current NN are identified; fourth, if worth-learning data are identified, these data are added to the training set and returns to the second step; otherwise, the current NN is output.

The above training process converges until no worth-learning data are identified. This means that the training set is sufficiently representative of the input space of the OPF problem. As a result, the NN trained based on this dataset can generalize to the input feasible set well. The following subsections introduce the proposed method in detail. 

\subsection{Physical-model-integrated NN}
\begin{figure}[bt]
    \centering
    \includegraphics[width=3.5in]{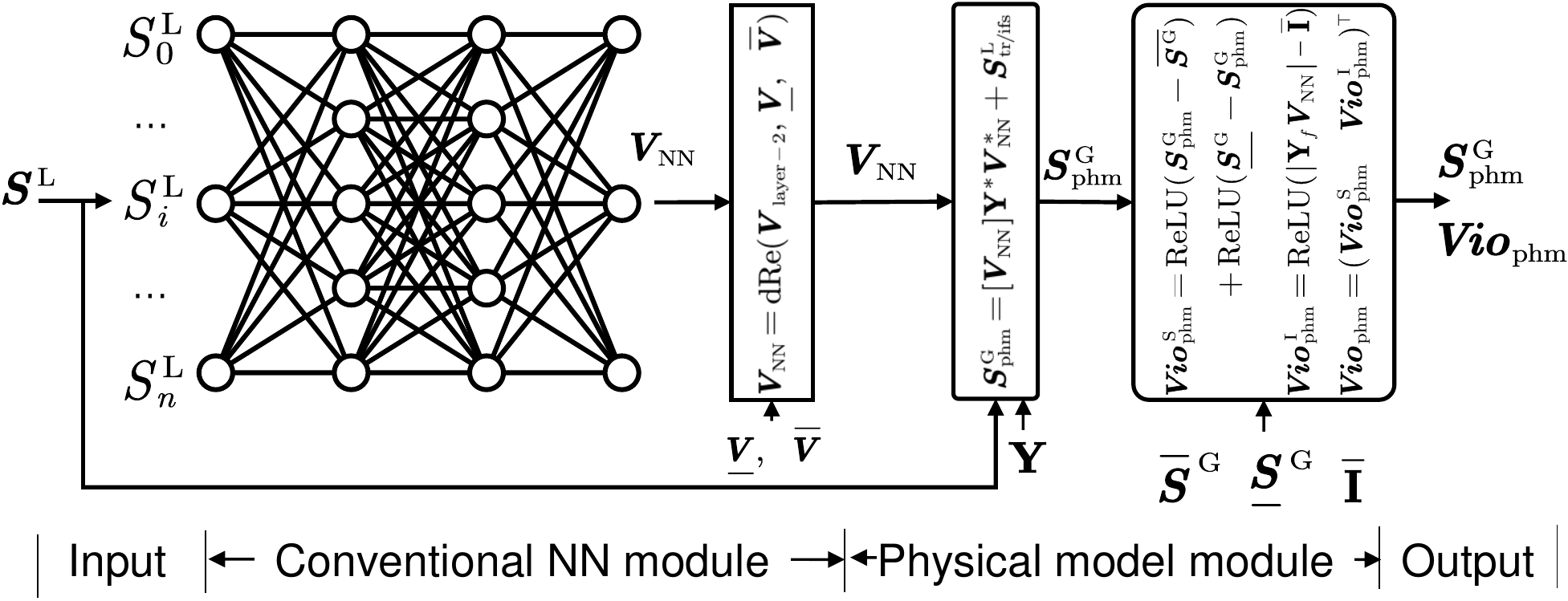}
    \vspace{-3mm}
    \caption{{The physical-model-integrated NN.}}
    \label{firsSecon} \vspace{-5mm}
\end{figure}

In the second step of the proposed method (\cref{trainflowchart}), the NN is trained to fit the mapping $ \boldsymbol{S}^{\text{G *}}=f^\text{OPF}(\boldsymbol{S}^{\text{L}})$. To obtain better results, we design a physical-model-integrated NN structure consisting of a \textit{conventional NN module} and a \textit{physical-model module}, as shown in \cref{firsSecon}. The former is a conventional MLP, while the latter is a computational graph transformed from the OPF model.
\subsubsection{Conventional NN module}
\begin{figure}[b] \vspace{-5mm}
% \vspace{-6mm}
    \centering
    \includegraphics[width=1.9in]{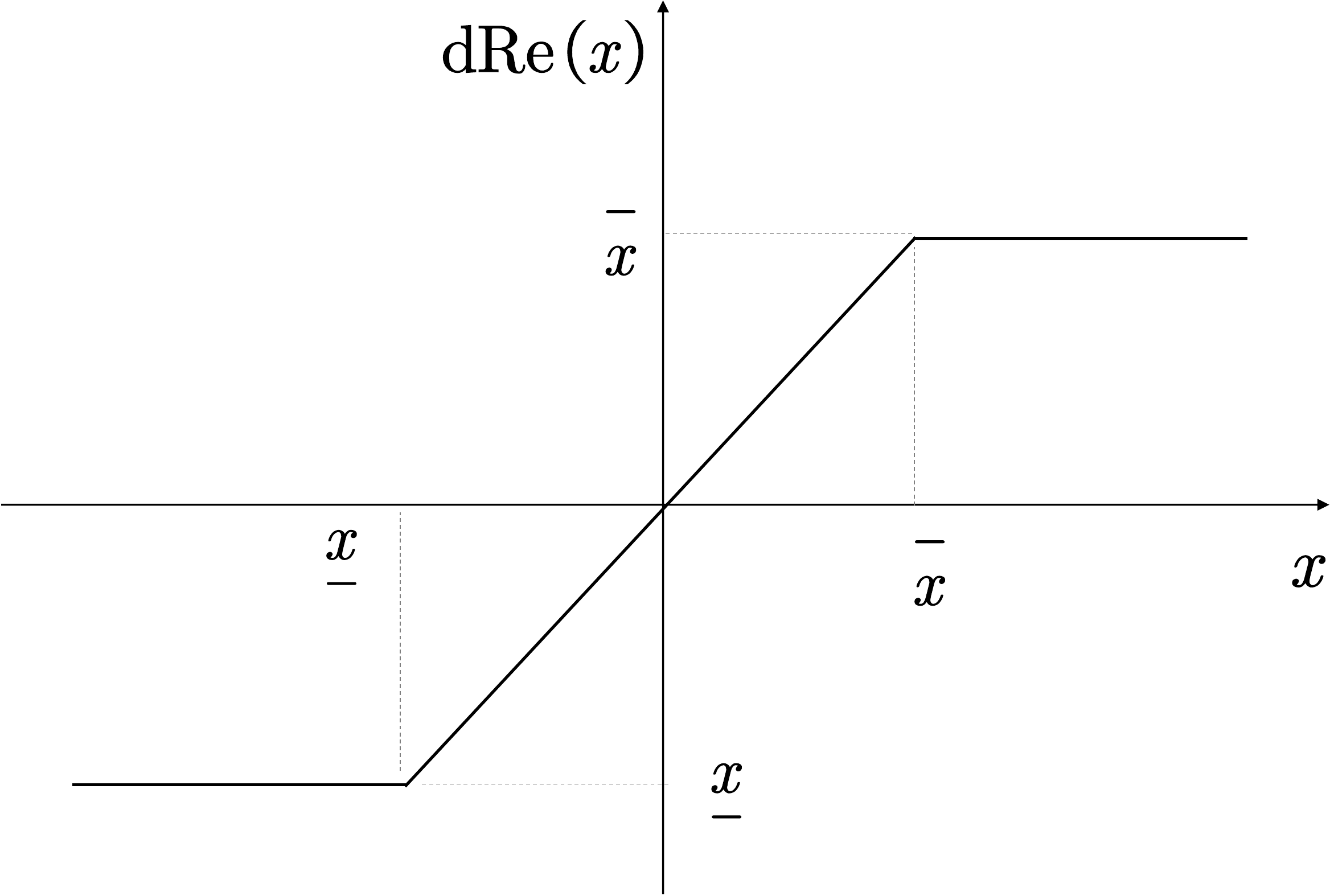}
    \caption{The dRe() function.}
    \label{dRePic} 
    \vspace{-2mm}
\end{figure}

This module first adopts a conventional MLP with learnable parameters to fit the mapping from the $\boldsymbol{S}^{\text{L}}$ to the optimal decision variable $\boldsymbol{V}_\text{NN}$ \cite{lowConvexRelaxationOptimal2014b}. The $\boldsymbol{V}_\text{NN}$ has its box constraint defined in \cref{AC-OPF model}. To ensure that the output $\boldsymbol{V}_\text{NN}$ satisfies this constraint, we design a function $\text{dRe}()$ to adjust any infeasible output $\boldsymbol{V}_\text{NN}$ into its feasible region, which is formulated as follows:
\begin{equation}\vspace{-0mm}
\begin{array}{ll}
   x \leftarrow \text{dRe}(x, \underline{x}, \overline{x})= \text{ReLU}(x-\underline{x})-\text{ReLU}(x-\overline{x})+\underline{x},& \label{dRe} 
\end{array}
\end{equation}
where $\text{ReLU(x)}=\text{max}(x,0)$;  $x$ is the input of the function, and its lower and upper bounds are $\underline{x}$ and $\overline{x}$, respectively. The diagram of this function is illustrated in \cref{dRePic}.

Applying $\text{dRe}()$ as the activation function of the last layer of the conventional MLP, the mathematical model of this conventional NN module is formulated as follows: 
\begin{align}
    &\boldsymbol{V}_\text{NN} = \text{MLP}(\boldsymbol{S}^{\text{L}}),\label{MLP_1}\\
    &\boldsymbol{V}_\text{NN} \leftarrow \text{dRe}(\boldsymbol{V}_\text{NN}, \underline{\boldsymbol{V}}, \overline{\boldsymbol{V}}),\label{MLP_2}
\end{align}
where \cref{MLP_1} describes the conventional model of MLP and \cref{MLP_2} adjusts the output of the MLP.

\subsubsection{Physical model module}

This module receives $\boldsymbol{V}_\text{NN}$ from the previous module, and then it outputs the optimal power generation $\boldsymbol{S}^{\text{G}}_\text{phm}$ and the corresponding constraints violation $\boldsymbol{Vio}_\text{phm}$, where the subscript ``$\text{phm}$" denotes the \textit{physical model module}. The first output $\boldsymbol{S}^{\text{G}}_\text{phm}$ is the optimal decision variable of the AC-OPF problem. It can be calculated by $\boldsymbol{V}_\text{NN}$ and $\boldsymbol{S}^{\text{L}}$, as follows:
\begin{align} \vspace{-0mm}
    \boldsymbol{S}^{\text{G}}_\text{phm}=[\boldsymbol{V}_{\text{NN}}]\textbf{Y}^* \boldsymbol{V}_{\text{NN}}^{*}+\boldsymbol{S}^{\text{L}}. \label{equ:phm1}
\end{align} \vspace{-0mm}
The second output $\boldsymbol{Vio}_\text{phm}$ (termed as violation degree) measures the quality of $\boldsymbol{S}^{\text{G}}_\text{phm}$ and is the key metric to guide the proposed worth-learning data generation (see details in the following subsection \ref{Sec_worthdata}). Given $\boldsymbol{V}_\text{NN}$ and $\boldsymbol{S}^{\text{G}}_\text{phm}$,  the violations of inequality constraints of the AC-OPF problem $\boldsymbol{Vio}_\text{phm}$ are calculated as follows:
\begin{subequations} \label{equ:phm2}
\begin{align} 
    &\boldsymbol{Vio}_\text{phm}^\text{S} = \text{ReLU}(\boldsymbol{S}^\text{G}_\text{phm}-\overline{\boldsymbol{S}}^\text{G})+ \text{ReLU}(\underline{\boldsymbol{S}}^\text{G}-\boldsymbol{S}^\text{G}_\text{phm}),\\
    &\boldsymbol{Vio}_\text{phm}^\text{I}= \text{ReLU}(|\textbf{Y}_{f}\boldsymbol{V}_\text{NN}|-\mathbf{\bar{I}}),\\
    & \boldsymbol{Vio}_\text{phm} = ( \boldsymbol{Vio}_\text{phm}^\text{S} \quad \boldsymbol{Vio}_\text{phm}^\text{I})^{\top},
\end{align}
\end{subequations} 
where $\boldsymbol{Vio}_\text{phm}^\text{S}$ denotes the violation of the upper or lower limit of $\boldsymbol{S}^\text{G}_\text{phm}$, and $\boldsymbol{Vio}_\text{phm}^\text{I}$ represents the violation of branch currents.

% By concatenating \boldsymbol{Vio}_\text{phm}^\text{S}\boldsymbol{Vio}_\text{phm}^\text{S} and \boldsymbol{Vio}_\text{phm}^\text{I}\boldsymbol{Vio}_\text{phm}^\text{I}, we define \boldsymbol{Vio}_\text{phm}\boldsymbol{Vio}_\text{phm}s the violation degree of  \boldsymbol{S}^\text{G}_\text{phm}\boldsymbol{S}^\text{G}_\text{phm}, i.e., the key variable for worth-learning dasta generation:
% \begin{equation}
%     \boldsymbol{Vio}_\text{phm} = ( \boldsymbol{Vio}_\text{phm}^\text{S} \quad \boldsymbol{Vio}_\text{phm}^\text{I})^{\top},
% \end{equation}

% \textcolor{red}{
\textit{Remark 1. The physical-model-integrated NN is formed by combining the \textit{conventional NN module} and the \textit{physical model module}. It inputs $\boldsymbol{S}^{\text{L}}$ and outputs $\boldsymbol{S}^{\text{G}}_\text{phm}$ and $\boldsymbol{Vio}_\text{phm}$, as shown in \cref{firsSecon}. Its function is the same as conventional OPF numerical solvers. In addition, it is convenient for users to directly determine whether the result of the NN OPF solver is acceptable or not based on the violation degree $\boldsymbol{Vio}_\text{phm}$. In contrast, most NN OPF solvers in the literature are incapable of outputting the violation degree directly \cite{OPFlarg, Deep0opf, sen-inf}.}

\subsubsection{Loss function}
\label{lossFunc}
To enhance the training accuracy of the physical-model-integrated NN, we design an elaborate loss function, which consists of $\boldsymbol{V}_{\text{NN}}$ from the \textit{conventional NN module}, and $\boldsymbol{S}^\text{G}_\text{phm}$ and  $\boldsymbol{Vio}_\text{phm}$ from the \textit{physical model module}. The formula is as follows:
% \begin{subequations}
\begin{gather}
\label{loss}
    % Vio = [M_r \quad M_i \quad M_I \quad M_{bus}_\text{phm} ]^{\top}\\
    loss=||\hat{\boldsymbol{V}}-\boldsymbol{V}_{\text{NN}}||_1+||\hat{\boldsymbol{S}^\text{G}}-\boldsymbol{S}^\text{G}_\text{phm}||_1
    +{\boldsymbol{Vio}_\text{phm}},
\end{gather}
where $\hat{\boldsymbol{V}}$ and $\hat{\boldsymbol{S}^\text{G}}$ are label values from the training set, which is a ground truth dataset from numerical solvers. 

\begin{figure}[tb]
    \centering
    \includegraphics[width=3.3in]{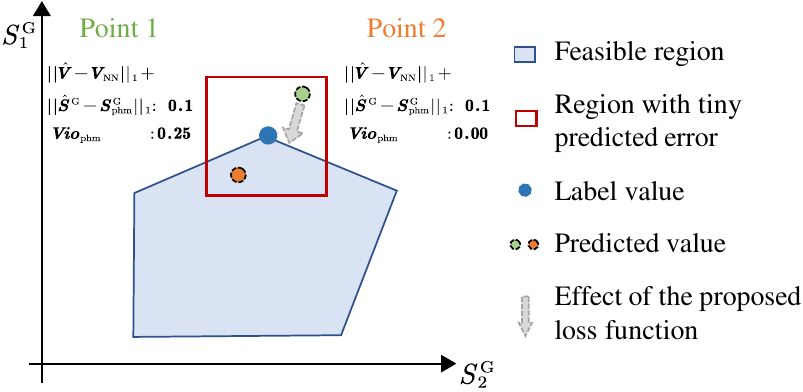}
    \caption{Illustration of the effectiveness of the three terms in the loss function.}
    \label{fig:lossThree} 
\end{figure}

{Combining the three terms in the loss function can help enhance fitting precision. As shown in \cref{fig:lossThree}, if the loss function only has the first two items $||\hat{\boldsymbol{V}}-\boldsymbol{V}_{\text{NN}}||_1$+ $||\hat{\boldsymbol{S}^\text{G}}-\boldsymbol{S}^\text{G}_\text{phm}||_1$ to penalize conventional fitting errors, the predicted value will be in a tiny square space (the red square in \cref{fig:lossThree}) around the label value. 
From the optimization perspective, the optimal label value is usually on the edge of its feasible region (the blue polyhedron in \cref{fig:lossThree}). This edge through the label value splits the square into two parts: the feasible (blue) part and the infeasible (white) part.  
Intuitively, we would prefer the predicted values to be in the feasible part. Thus, we also penalize violation degree $\boldsymbol{Vio}_\text{phm}$ in the loss function to force the predicted values with big $\boldsymbol{Vio}_\text{phm}$ close to the square's feasible half space for smaller constraint violations.}

Although the proposed NN with elaborate loss function has high training accuracy, it is still difficult to guarantee the generalization of the NN OPF solver to the whole input space with conventional random sampling. Therefore, it is indispensable and challenging to obtain a representative training dataset with moderate size to train the proposed NN, which is the focus of the following subsection. 

\subsection{Worth-learning data generation}\label{Sec_worthdata}
As shown in \cref{trainflowchart}, we adopt an iterative process to identify the worth-learning data. For an NN trained in the previous iteration, we utilize its output $\boldsymbol{Vio}_\text{phm}$ to help identify new data samples that are not yet properly generalized. Specifically, if an input $\boldsymbol{S}^\text{L*}$ is feasible for the original OPF problem while the current NN outputs a large violation degree $\boldsymbol{Vio}_\text{phm}^*$, the contradiction means the NN has a large fitting error at $\boldsymbol{S}^\text{L*}$. This is probably because {sample $\boldsymbol{S}^\text{L*}$ was not included in the previous training set and was not generalized by the NN}. Hence, this sample $\boldsymbol{S}^\text{L*}$ can be regarded as a \textit{worth-learning sample}. Including the sample in the training dataset in the next iteration helps enhance the generalization of the NN. 

% \textcolor{red}{To obtain a representative training set, each data generated should be worth-learning for the NN, which are the third and fourth steps of the proposed method (\cref{trainflowchart}). Specifically, in the iterative process in \cref{trainflowchart}, identifying data and training NNs are conducted alternately. i.e., after one training step, we identify some feasible inputs generalized badly by the current NN and then patch the NN's limitations with the new dataset in the next training step.}  

The key to the proposed worth-learning data generation method is to identify worth-learning samples efficiently. Instead of traversing all of the possible inputs, we maximize $\boldsymbol{Vio}_\text{phm}$ for a given NN to identify the input with a large violation degree. However, the inputs identified in the maximizing process should be feasible for the original OPF problem. Otherwise, the found inputs might be infeasible and useless for the representation of the training data.
% design a max violation backpropagation method (\cref{MAxviosubsection}) to identify those inputs.  
% that 
% If the identified inputs lead to large violation degrees \textcolor{green}{and are feasible for the original OPF problem},  they are \textcolor{blue}{also} the worth-learning samples for the NN and shall be included in the training set.

\subsubsection{Input feasible set module} 
To keep the inputs identified in the maximizing process feasible for the original OPF problem, we formulate the \textit{input feasible set module} to restrict power loads $\boldsymbol{S}^\text{L}$ to their feasible set.
The \textit{feasible set} is composed of box constraints, current limits, and KCL\&KVL constraints, which  are transformed from the feasible set of the OPF problem defined in \cref{AC-OPF model}.
The partial formulations of the \textit{input feasible set} are as follows, where the subscript ``$\text{ifs}$" denotes the \textit{input feasible set module}:
\begin{subequations}
\vspace{-1mm}
\label{fronlayer1}
\begin{align}
    \boldsymbol{S}^\text{G}_\text{ifs}& =\text{dRe}\left( \boldsymbol{S'}^\text{G}_\text{ifs},\,\underline{\boldsymbol{S}}^\text{G},\,\overline{\boldsymbol{S}}^\text{G} \right),\,\, \boldsymbol{S'}^\text{G}_\text{ifs}\in\mathbb{R} ^n, \label{S^G} \\
    \boldsymbol{V}_\text{ifs}& =\text{dRe}\left( \boldsymbol{V'}_\text{ifs},\,\underline{\boldsymbol{V}},\,\overline{\boldsymbol{V}} \right),\,\,  \boldsymbol{V'}_\text{ifs}\in \mathbb{R} ^n, \label{Vfron}\\ 
    \boldsymbol{S}^\text{L}_\text{ifs}& =\boldsymbol{S}^\text{G}_\text{ifs}\,\,-\,\,[\boldsymbol{V}_\text{ifs}]\textbf{Y}^*\boldsymbol{V}^*_\text{ifs} ,\label{CVLinFron}	\\
    \boldsymbol{I}_\text{ifs}& = \mathbf{Y}_\text{b}\boldsymbol{V}_\text{ifs} ,\label{CurrentinFron}
\end{align}
\end{subequations}
where $\boldsymbol{S'}^\text{G}_\text{ifs}$ and  $\boldsymbol{V'}_\text{ifs}$ are auxiliary $n\times 1$ vectors in $\mathbb{R} ^n$ and have no physical meaning. Symbols $\boldsymbol{S}^\text{G}_\text{ifs}$ and $\boldsymbol{V}_\text{ifs}$ are restricted in their box constraints in \cref{S^G,Vfron}. Then the KCL\&KVL correlations of $\boldsymbol{S}^\text{L}_\text{ifs}$, $\boldsymbol{S}^\text{G}_\text{ifs}$, and $\boldsymbol{V}_\text{ifs}$ are described by \cref{CVLinFron}. Symbol $\boldsymbol{I}_\text{ifs}$ in \cref{CurrentinFron} denotes the currents at all branches. 

{The other formulations of the \textit{input feasible set} aim to calculate $\boldsymbol{Vio}_\text{ifs}$, the AC-OPF's constraint violations corresponding to $\boldsymbol{S}^\text{L}_\text{ifs}$ and $\boldsymbol{I}_\text{ifs}$, as follows: }
% {from box limits of  \boldsymbol{S}^\text{L}_\text{ifs}\boldsymbol{S}^\text{L}_\text{ifs} and \boldsymbol{I}_\text{ifs}\boldsymbol{I}_\text{ifs}. Instead of the \text{dRe}()\text{dRe}() function, symbols \boldsymbol{Vio}^\text{S}_\text{ifs}\boldsymbol{Vio}^\text{S}_\text{ifs} and   \boldsymbol{Vio}^\text{I}_\text{ifs}\boldsymbol{Vio}^\text{I}_\text{ifs} are used to describe box constraints of \boldsymbol{S}^\text{L}_\text{ifs}\boldsymbol{S}^\text{L}_\text{ifs}, \boldsymbol{I}_\text{ifs}\boldsymbol{I}_\text{ifs},   because they are also determined by other variables in \cref{CVLinFron,CurrentinFron}, cutting the violations directly by the \text{dRe}()\text{dRe}() would lead to the inequalities of  \cref{CVLinFron,CurrentinFron}.  The  equations are: }
\begin{subequations}
\label{frn1cg}
\begin{align}
\boldsymbol{Vio}^\text{S}_\text{ifs}& = \text{ReLU}(\boldsymbol{S}^\text{L}_\text{ifs}-\overline{\boldsymbol{S}}^\text{L})+ \text{ReLU}(\underline{\boldsymbol{S}}^\text{L}-\boldsymbol{S}^\text{L}_\text{ifs}),\label{S^L}\\
\boldsymbol{Vio}^\text{I}_\text{ifs}& = \text{ReLU}(\boldsymbol{|I_\text{ifs}|}-\overline{\mathbf{I}}),\\
\boldsymbol{Vio}_\text{ifs}& = ( \boldsymbol{Vio}_\text{ifs}^\text{S} \quad \boldsymbol{Vio}_\text{ifs}^\text{I} )^{\top},\label{currentFron}
\end{align}
\end{subequations}
where $\boldsymbol{Vio}_\text{ifs}^\text{S}$ denotes the violation of the upper or lower limit of $\boldsymbol{S}^\text{L}_\text{phm}$, and $\boldsymbol{Vio}_\text{ifs}^\text{I}$ denotes the violation of branch current.
% \boldsymbol{Vio}_\text{ifs}\boldsymbol{Vio}_\text{ifs} denotes the constraint violation of the \textit{input feasible set module}.
% The \textit{input feasible set module} is constructed by the  box constraints \cref{S^G,Vfron,S^L}, KCL\&KVL correlation \cref{CVLinFron}, and current limits \cref{CurrentinFron,currentFron}, as shown in \cref{inputFea}.
\begin{figure}[bt]\vspace{-2mm}
    \centering
    \includegraphics[width=2.3in]{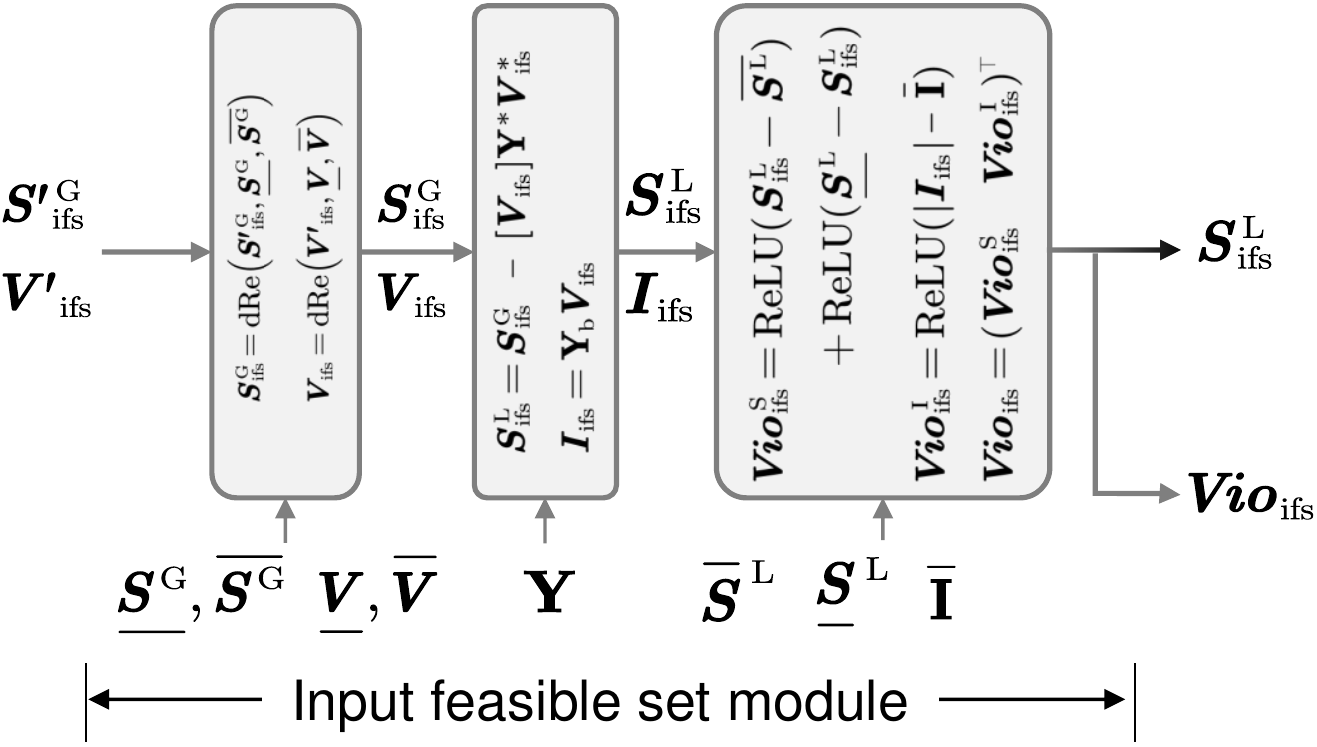}
    \caption{The \textit{input feasible set module}.}
    \label{inputFea} \vspace{-4mm}
\end{figure}
\textit{Remark 2. This module takes $\boldsymbol{S'}^\text{G}_\text{ifs}$ and $\boldsymbol{V'}_\text{ifs}$ as the inputs, and then outputs $\boldsymbol{S}^\text{L}_\text{ifs}$ and $\boldsymbol{Vio}_\text{ifs}$, as shown in \cref{inputFea}. When $\boldsymbol{Vio}_\text{ifs}= 0$, the corresponding $\boldsymbol{S}^\text{L}_\text{ifs}$ lies in the feasible set of the AC-OPF problem. To identify feasible $\boldsymbol{S}^\text{L}_\text{ifs}$ in the process of maximizing $\boldsymbol{Vio}_\text{phm}$, this module backpropagate the $\frac{\partial\boldsymbol{Vio}_\text{phm}}{\partial\boldsymbol{S}^\text{L}_\text{ifs}}$ with $\boldsymbol{Vio}_\text{ifs}\leq \zeta$ ($\zeta$ is a small positive tolerance), and then it updates $\boldsymbol{S'}^\text{G}_\text{ifs}$ and $\boldsymbol{V'}_\text{ifs}$. As a result, the corresponding $\boldsymbol{S}^\text{L}_\text{ifs}$ is always feasible. Furthermore, because $\boldsymbol{S'}^\text{G}_\text{ifs}$ and $\boldsymbol{V'}_\text{ifs}$ are not bounded, changing them can theoretically find any feasible $\boldsymbol{S}^\text{L}_\text{ifs}$.}

\subsubsection{Max violation backpropagation}
\label{MAxviosubsection}
\begin{figure}[bt] 
    \centering
    \includegraphics[width=3.0in]{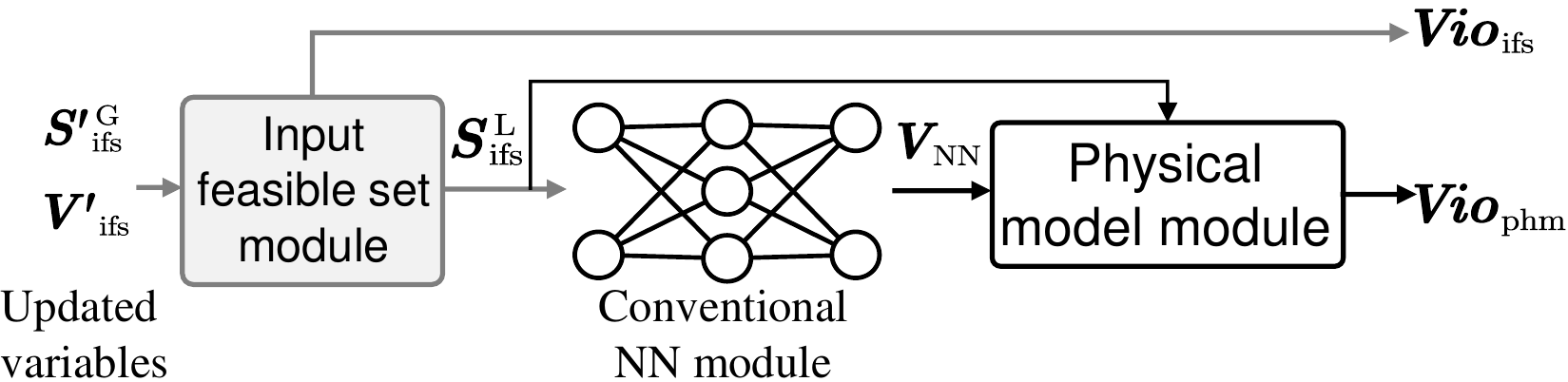}
    \caption{The novel NN for max violation backpropagation by integrating physical-model-integrated NN with the \textit{input feasible set module}.} 
    \label{theMP4Label}
    \vspace{-2mm}
\end{figure}

To identify worth-learning data, a novel NN is created by inputting $\boldsymbol{S}^\text{L}_\text{ifs}$ into the physical-model-integrated NN (see \cref{theMP4Label}). This NN has two outputs, i.e., $\boldsymbol{Vio}_\text{phm}$ and  $\boldsymbol{Vio}_\text{ifs}$. The former measures the constraint violation degree of the OPF solution $\boldsymbol{S}^{\text{G*}}$; the latter indicates the feasibility of the OPF input $\boldsymbol{S}^\text{L}_\text{ifs}$. If $\boldsymbol{S}^\text{L}_\text{ifs}$ is a feasible input, i.e., $\boldsymbol{Vio}_\text{ifs}\leq \zeta$, but the optimal solution $\boldsymbol{S}^{\text{G*}}$ is infeasible, i.e., $\boldsymbol{Vio}_\text{phm}\geq \xi$ {($\xi$ is a threshold)}, this means the corresponding input is worth learning (i.e., it is not learned or generalized by the current NN). Based on this analysis, we design the loss function $loss_\text{max}$ for max violation backpropagation, as follows: 
\begin{align}
    loss_{\text{max}} = \boldsymbol{Vio}_\text{phm} - \lambda \times \boldsymbol{Vio}_\text{ifs} \label{maxLoss},
\end{align}
where $\lambda$ is a large, constant weight parameter. When maximizing this loss function, the algorithm tends to find a worth-learning $\boldsymbol{S}^\text{L}_\text{ifs}$ that has small $\boldsymbol{Vio}_\text{ifs}$ but large $\boldsymbol{Vio}_\text{phm}$. 
% The  maximized $loss_\text{max}$ denotes a large violation degree for $\boldsymbol{S}^\text{L}_\text{ifs}$ and the $\boldsymbol{S}^\text{L}_\text{ifs}$ is feasible or close to feasible with large $\boldsymbol{Vio}_\text{phm}$ and large $-\lambda \times \boldsymbol{Vio}_\text{ifs}$, respectively, so this $\boldsymbol{S}^\text{L}_\text{ifs}$ is the worth-learning data. 

{During the max violation backpropagation, the proposed algorithm maximizes $loss_\text{max}$ to update the variables $\boldsymbol{S'}^\text{G}_\text{ifs}$ and $\boldsymbol{V'}_\text{ifs}$ by gradient backpropagation until $loss_\text{max}$ converges to the local maximum. After the process, the corresponding $\boldsymbol{S}^\text{L}_\text{ifs}$ is also found.}
{Because the maximizing process can be processed in parallel by the deep learning module PyTorch, the worth-learning samples are found in batch, where the max violation backpropagation uses the previous training set as initial points to identify the new data.} Further, the auto-differentiation technique in PyTorch can accelerate the process of parallel computation. Based on these techniques, massive worth-learning data samples are identified efficiently.

\subsection{Overall training process}
The overall training process is presented in \cref{alg:the_alg}, which first takes an initial training dataset $\mathbb{D}_\text{t}$ (obtained by any conventional sampling method) as input. The learning rate $\eta$ is equal to $10^{-3}$, the loss difference tolerance $\epsilon$  is equal to $10^{-2}$, the added dataset $\mathbb{A}$ is empty, and the loss difference $\Delta L$  is equal to infinity at initialization. The training is performed for a fixed number of epochs (lines 2--5). Then the max violation backpropagation starts to identify worth-learning data (lines 6 and 7) by using the training data as the initial points (line 8) and updating $\boldsymbol{S'}^\text{G}_\text{ifs}$ and $\boldsymbol{V'}_\text{ifs}$ until $\Delta L$ is less than $\epsilon$ (lines 9--12), which indicates $loss_{max}$ has converged to the terminal.

{After the max violation backpropagation, a series of commands are designed to add proper data to the training set. First, a filter function $f_\text{filter}$ is employed to eliminate data with terminal violation $\boldsymbol{Vio}_{\text{phm}, N}$ less than a given threshold $\xi$ (the value depends on the acceptable violation settings). Second, $ \{ \hat{\boldsymbol{V}},\hat{\boldsymbol{S}^\text{G}} \}$ is calculated by numerical solvers corresponding to $\boldsymbol{S}^\text{L}_\text{ifs}$ with large violation degree (lines 14 and 15). They consist of added set $\mathbb{A}$ (line 16). Third, the training set $\mathbb{D}_t$ is expanded with $\mathbb{A}$ (line 17). The loop is repeated until the added set $\mathbb{A}$ is empty (line 18), meaning no worth-learning data are identified.}

\begin{algorithm}[bt]
 \caption{Training process of the physical-model-integrated NN OPF solver with worth-learning data generation.}
 \label{alg:the_alg}
 \begin{small}
 \begin{algorithmic}[1]
 \renewcommand{\algorithmicrequire}{\textbf{Input:}}
 \renewcommand{\algorithmicensure}{\textbf{Output:}}
 \REQUIRE $\mathbb{D}_\text{t} =\left(\hat{\boldsymbol{S}^\text{L}},\hat{\boldsymbol{V}},\hat{\boldsymbol{S}^\text{G}} \right)$ \\
 \textit{Initialization} : $\eta \gets 10^{-3}, \epsilon \gets 10^{-2}, \mathbb{A}\gets  \varnothing, \Delta L \gets \infty$\\
        \REPEAT
        \FOR {$\text{epoch}$ $k=0,1,...$}
        \STATE Train the NN with $loss$ \cref{loss}:
        \STATE $w \gets w-\eta \nabla loss$.
        \ENDFOR
        \WHILE{$\Delta L \ge \epsilon$} 
        \STATE Identify data with $loss_\text{max}$ \cref{maxLoss}:
        \STATE $\boldsymbol{S'}^\text{G}_\text{ifs}$, $\boldsymbol{V'}_\text{ifs} \gets$  $\boldsymbol{S}^\text{G}_\text{ifs},\,\, \boldsymbol{V}_\text{ifs} \gets \hat{\boldsymbol{S}^\text{G}},\,\, \hat{\boldsymbol{V}}$
        \STATE $\boldsymbol{S'}^\text{G}_\text{ifs} \gets \boldsymbol{S'}^\text{G}_\text{ifs} + \eta \nabla loss_\text{max}$
        \STATE $\boldsymbol{V'}_\text{ifs} \gets \boldsymbol{V'}_\text{ifs} + \eta \nabla loss_\text{max}$
        \STATE $\Delta L \gets |\,loss_{\text{max},i} - loss_{\text{max},i-100}\,|   $
        \ENDWHILE
        \STATE $\{\boldsymbol{Vio}_{\text{phm},N}\} \gets f_\text{filter}(\boldsymbol{Vio}_{\text{phm},N}\ge \xi)$
        \STATE Collect $\{ \boldsymbol{S}^\text{L}_\text{ifs} \}$ corresponding to $\{\boldsymbol{Vio}_{\text{phm},N}\}$ based on the novel NN in \cref{theMP4Label}
        \STATE Calculate $ \{ \hat{\boldsymbol{V}},\hat{\boldsymbol{S}^\text{G}} \} $ corresponding to $\{\boldsymbol{S}^\text{G}_\text{ifs}\}$ using numerical solvers
        \STATE $\mathbb{A}\gets \{\boldsymbol{S}^\text{L}_\text{ifs}, \hat{\boldsymbol{V}}, \hat{\boldsymbol{S}^\text{G}}\} $
        \STATE $\mathbb{D}_\text{t}\gets \mathbb{D}_\text{t} \cup \mathbb{A}$
		\UNTIL $\mathbb{A}$ is  $\varnothing$
 \end{algorithmic} 
 \end{small}
 \end{algorithm}

\subsection{Efficiency and convergence of the proposed method}
\label{Section 4}
Unlike general training processes for conventional NNs, the proposed physical-model-integrated NN with worth-learning data generation adopts an iterative training process. It iteratively checks the NN's generalization to the input's feasible space by identifying worth-learning data, as shown in \cref{trainflowchart} and \cref{alg:the_alg}.
{This difference introduces two critical questions. 1) Efficiency: is the process of identifying worth-learning data computationally efficient? 2) Convergence: is the training set representative of the whole input space after iterations?}
In terms of the computational efficiency of the proposed method, the theoretical analysis (detailed in the Appendix \ref{appendix:efficiencyofMVB}) shows it takes no more than 0.08 s to find one sample, which brings little computational burden into the training process. According to the experiment results, the average consumption time for finding one sample is 0.056 s. In terms of the convergence, we prove that the training set would gradually represent the whole input space in the Appendix \ref{appendix:convergenceMVB}, because the number of worth-learning samples identified would converge to zero after a finite number of iterations. 

\section{Numerical Experiments}
\label{Section 5}
The proposed method is evaluated using the IEEE 12-bus, 14-bus, 30-bus, 57-bus, and 118-bus systems. The ground truth datasets are constructed using PANDAPOWER based on a prime-dual interior points algorithm.
\subsection{The efficiency of worth-learning data generation}
\label{section:effi}
\begin{figure}[bt] %设置浮动属性，不设的话，图片可能会出现在页面的顶部，而不是在文字后面。
  \centering %居中
  \includegraphics[width=2.3in]{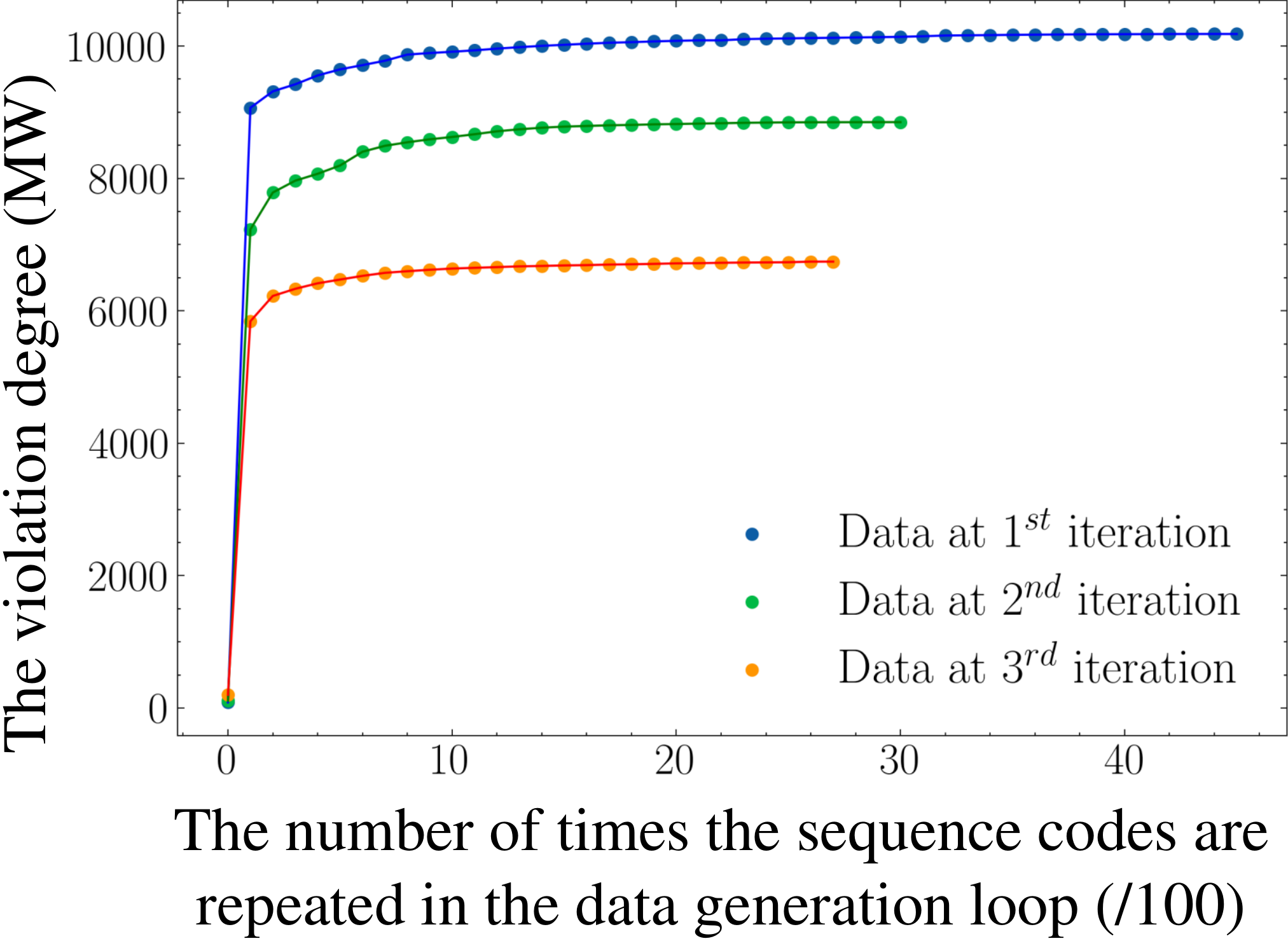} %设置宽度，一般a4纸是8inch。这里的单位也可以用cm等其它latex支持的。
  \caption{{Time consumption of the worth-learning data codes in three different iterations. The number of times the sequence codes are repeated in the data generation loop (x-axis) represents the time consumed in one data generation loop; the violation degrees (y-axis) quickly converge to the terminal stage.}} %caption是图片的标题
  \label{cal_efficiency} %此处的label相当于一个图片的专属标志，目的是方便上下文的引用。可以用\ref{img2}\ref{img2}来进行交叉引用，注意在引文中只是显示了一个第几张图片这个数字，还要自己在正文中补充其它内容。
  \vspace{-3mm}
\end{figure}

 As shown in \cref{alg:the_alg}, the proposed worth-learning data generation (lines 6--12) is the second loop in one iteration (lines 1--18), and the number of initial data points for the generation varies with iterations (lines 8, 15--17). To evaluate the efficiency of the worth-learning data generation, we conduct an experiment on the IEEE 57-bus system in three different iterations to quantitatively measure how much time it takes to finish one worth-learning data generation loop. The time consumption of the data-generation loops in the three different iterations is illustrated in \cref{cal_efficiency}. The x-axis is the number of times the codes are repeated (lines 6--12) divided by 100, which represents the time consumed in one data generation loop; the y-axis is the violation degree. The three lines converge to the terminal stage within 4000 {times}. The trends are similar: they increase very quickly at first (with 100 epochs) and then approach the local maximum slowly (with 2900--3900 epochs).  The inflection points on the three lines are $(1, 7228)$, $(1,9065)$, and $(1,5841)$.

% , whose x coordinate values are 1. That means the updated data is close to their local maximum with 100 epochs. Then, the remaining 3900 epochs are taken to approach the local maximum until the delta is less than the threshold. 

% would be identified by max violation backpropagation based on them. The maximizing process in one iteration is completed when the delta of violation degrees between n and n+1 steps is less than 10^{-2}10^{-2} . Moreover, the added data are filtered by comparing their violation degree with a threshold value. The first three iterations of generating data are recorded for analysis. 

% The three iterations are shown in Fig. \ref{cal_efficiency}\ref{cal_efficiency}. 

In the three iterations, 300, 500, and 800 new data samples are identified. Each data-generation loop in iterations takes 30 s on average to run 3000--4000 times. Hence, one worth-learning data sample costs $(30\times 3)/ (300+500+800)\approx 0.056$ s, which introduces little computational burden into the training process compared to the other steps in \cref{alg:the_alg}. For example, each label value calculated by numerical solvers costs around 1 s (line 14), {and the NN training on a dataset with 1100 samples costs around 600 s (lines 2--5).}
In conclusion, the numerical experiment verifies that the worth-learning data generation brings little computational burden to the training process.
\renewcommand\arraystretch{0.2}
\begin{table}[]
\small
\centering
\caption{{Training time based on the conventional simple random sampling and proposed worth-learning data generation}}
\begin{tabular}{p{1.5cm} p{2.7cm} p{2.5cm}  }
\toprule
Cases & Conventional (min.) & Proposed (min.)  \\ \midrule
30-bus  & 27.9  & 30.1   \\ \midrule
57-bus  & 79.8  & 85.5   \\ \midrule
118-bus & 174.1 & 181.2  \\ \bottomrule
\end{tabular}%
\label{tab1}\vspace{-5mm}
\end{table}

Furthermore, we list the time consumption comparison of the conventional and proposed training processes in \cref{tab1}, where the conventional training process uses simple random sampling in place of the data generation loop (lines 6--12) in \cref{alg:the_alg}. By comparing the time consumption of the two methods, we can conclude that the training time of the proposed method only increases by 4\%--8\%. Hence, these experiments validate that the proposed worth-learning data generation is computationally efficient.

\subsection{Reliability and optimality of the proposed solver }
\label{reNopt}
To validate the superiority of the proposed NN OPF solver (denoted by \textbf{Proposed NN}), we compare it with two benchmarks: 1) \textbf{B1 NN}, which adopts the conventional loss function and NN model (MLP) with a training dataset generated by simple random sampling; 2) \textbf{B2 NN}, which adopts the proposed loss function and physical-model-integrated model with a training dataset generated by simple random sampling. 

A particular test set different from the training datasets above is created to examine the effect of these models fairly. {The test set has 600 samples that are produced by uniformly sampling 200 points in $[80\%, 120\%]$ of the $\text{nominal value}$ of one load three times. The other loads in the three times are fixed at light $(80\%\times \text{nominal value})$, nominal $(100\%\times \text{nominal value})$, and heavy $(120\%\times \text{nominal value})$ load conditions. The load sampled has the largest nominal value to cover a big region of the input space.
 Based on these settings, the test set includes much ``unseen" data for those models.}

The reliability of the NN OPF solvers is evaluated by the constraint violation degrees on all test data. The optimality loss is evaluated by the relative error between predicted results and label values. For a fair comparison, the three methods all stop their training processes when the value of $|| \hat{\boldsymbol{V}}-\boldsymbol{V}_\text{NN}||_1$ is less than $2\times 10^{-4}$. In view of the iterative training process, the performance of the three solvers is studied with increasing training data, and the initial NNs are identical because they are trained on an initial dataset with $N$ samples. 

\begin{figure}[t]
\flushright
\subfigure[] { \label{30bus-vio:b}   % The sum violation of constraints on 30-bus system
\includegraphics[width=1\columnwidth]{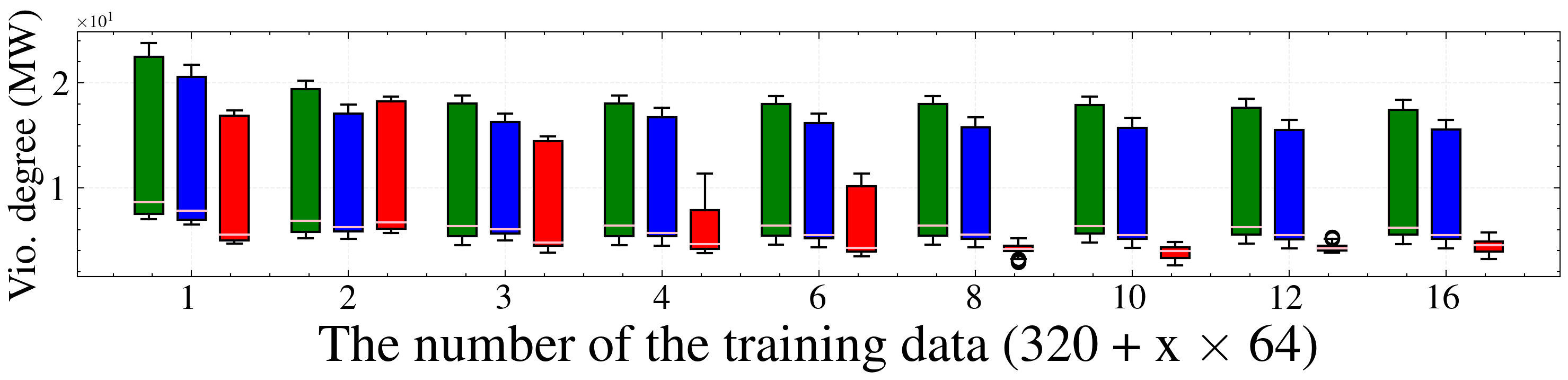}
}\vspace{-3mm}
\subfigure[] { \label{57bus-vio:b}     % The sum violation of constraints on 57-bus system
\includegraphics[width=1\columnwidth]{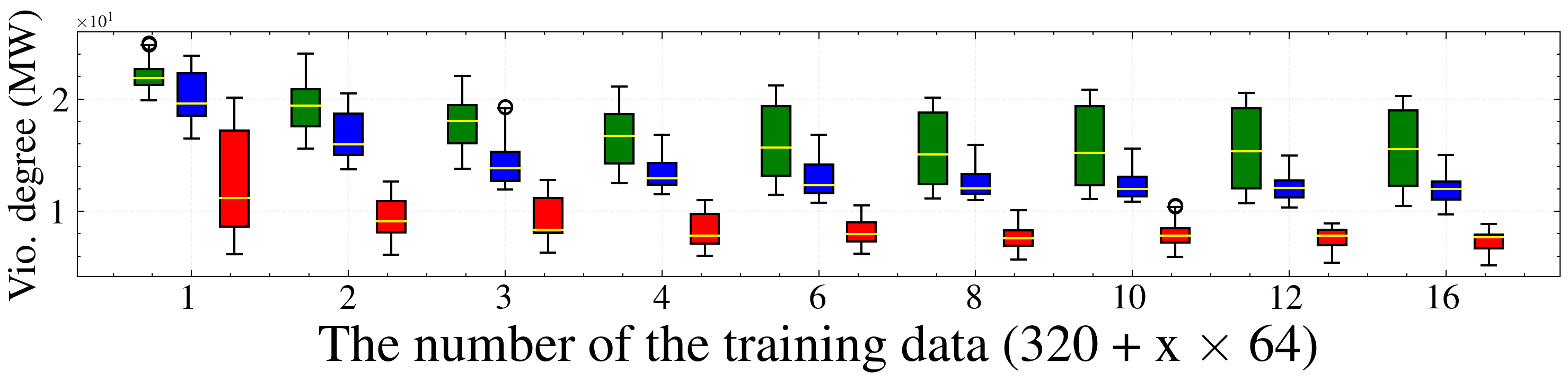}
}\vspace{-3mm}
\subfigure[] { \label{118bus-vio:b}      % The sum violation of constraints on 118-bus system
\includegraphics[width=1\columnwidth]{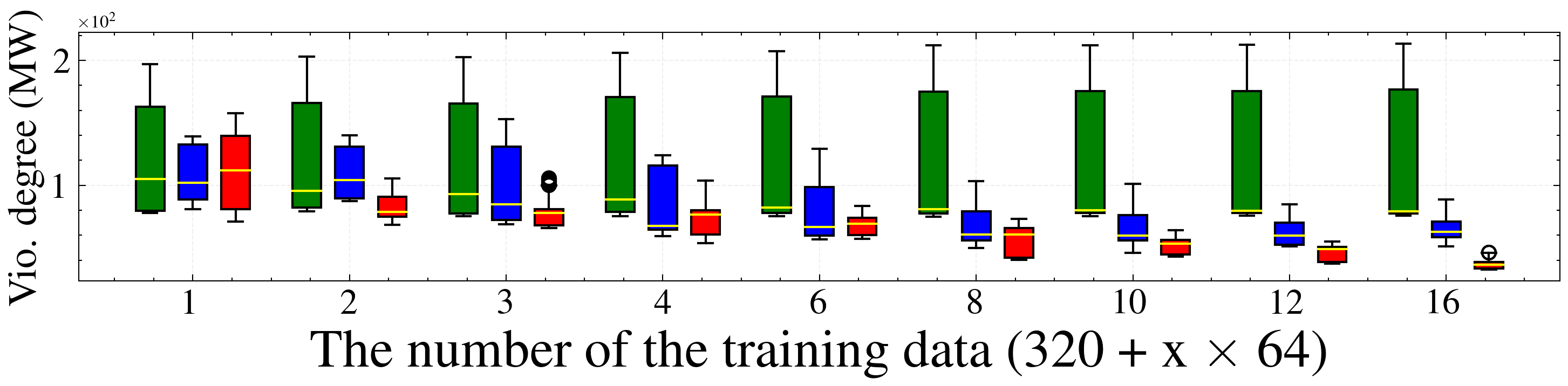}
}\vspace{-3mm}
\subfigure[] { \label{30bus-o:a}     % The optimality loss of predicted results on 30-bus system
\includegraphics[width=1\columnwidth]{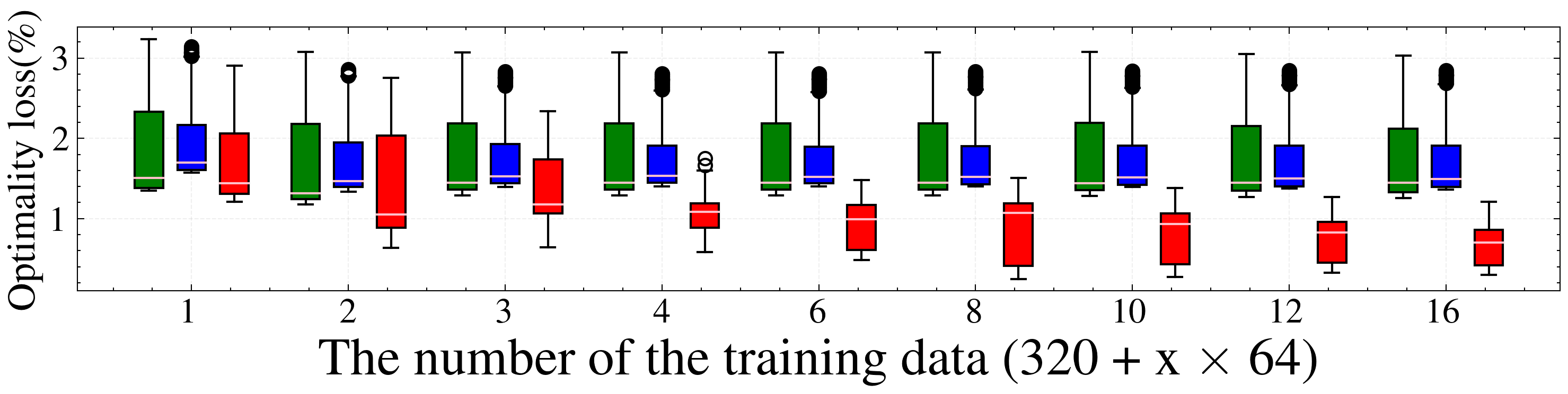}
}\vspace{-3mm}
\subfigure[] { \label{57bus-o:b}    % The optimality loss of predicted results on 57-bus system
\includegraphics[width=1\columnwidth]{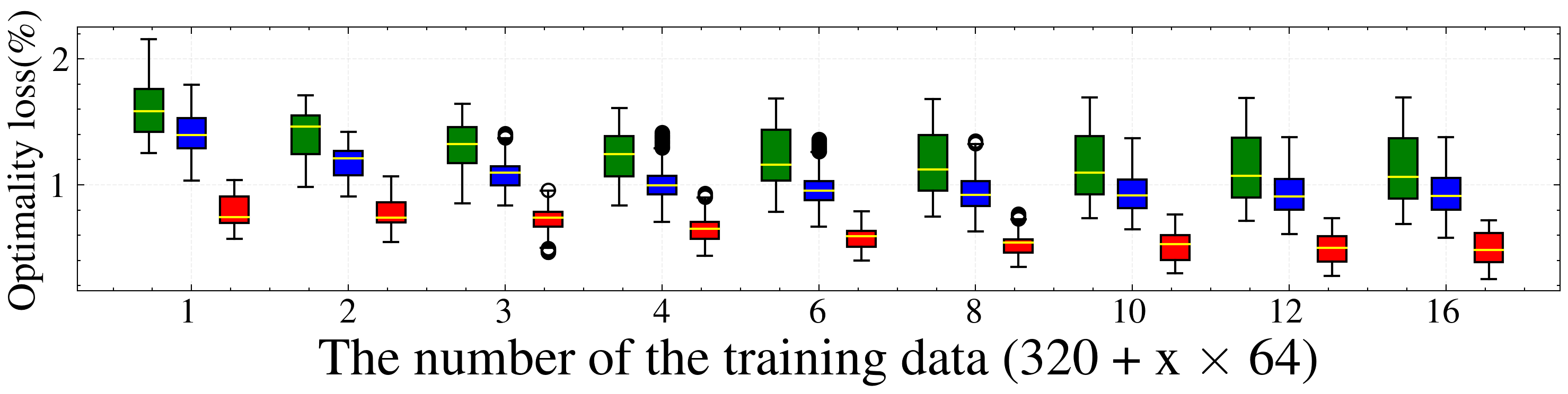}
}\vspace{-3mm}
\subfigure[]{ \vspace{-9mm} \label{118bus-o:b}
\includegraphics[width=1\columnwidth]{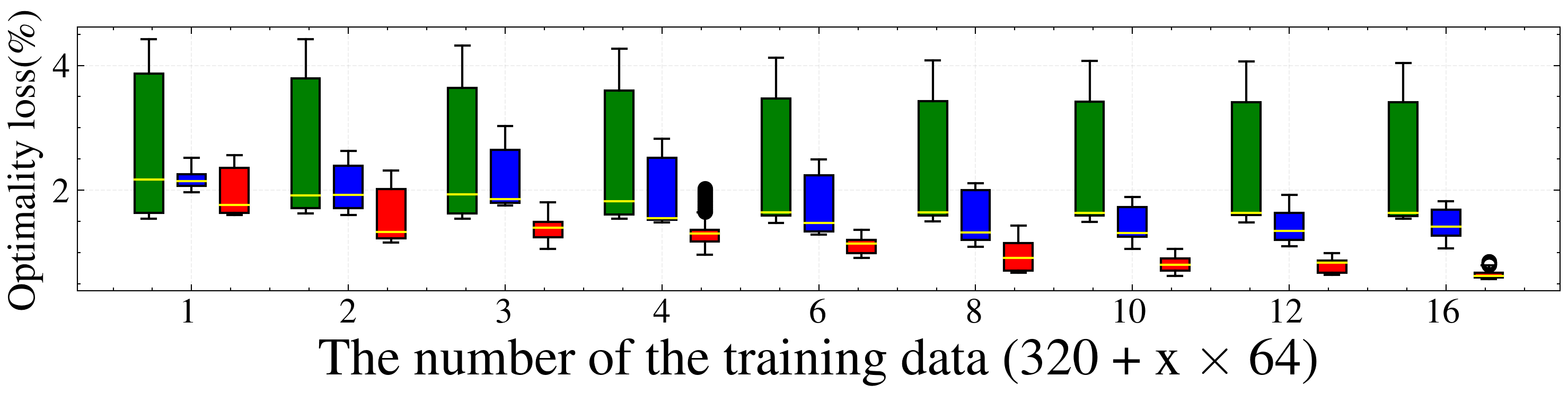}
}\vspace{-1mm}

\subfigure { 
\includegraphics[width=0.85\columnwidth]{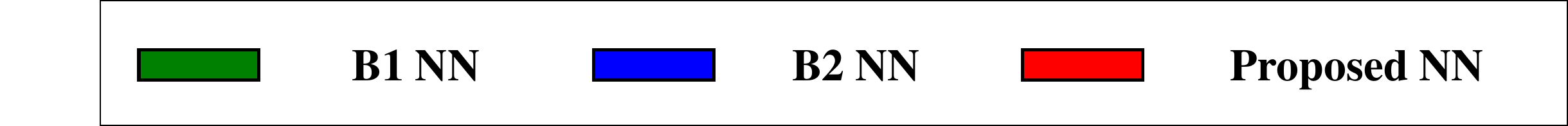}
}\vspace{-3mm}
\caption{The violation degree and optimality loss of the results of the NNs trained by three methods change with the number of training data in different cases: (a), (d) IEEE 30-bus; (b), (e) IEEE 57-bus; (c), (f) IEEE 118-bus. }\vspace{-5mm}
\label{opt_vio_3_cases}
\end{figure}
The results are statistically analyzed by creating box plots displayed in Fig. \ref{opt_vio_3_cases}. The violation degrees and optimality losses of the results of the NNs from the three methods converge to the terminal stages gradually. The rate of convergence of \textbf{Proposed NN} is the largest, that of \textbf{B2 NN} is in the middle, and that of \textbf{B1 NN} is the smallest.

In \cref{30bus-vio:b,57bus-vio:b,118bus-vio:b}, the comparison of the last violation degree gives notable results in the three cases. Specifically, the median values in three cases are 7, 15, and 75 for \textbf{B1 NN}; 6, 12.5, and 60 for \textbf{B2 NN}; and 3.2, 6.1, and 25 for \textbf{Proposed NN}, respectively. The novel loss function brings a 19\% reduction of violation degree on average by comparing \textbf{B1 NN} and \textbf{B2 NN}. The proposed training data generation method introduces a 50\% reduction of violation degree on average according to the comparison of \textbf{B2 NN} and \textbf{Proposed NN}. {Moreover, the height of the last boxes in each subfigure suggests the robustness of the three solvers, and \textbf{Proposed NN} has the smallest height in all three cases, which indicates the worth-learning data generation can improve the reliability in encountering ``unseen" data from the feasible region.}

% 56\%, 100\%, 220\%,
% and 60; the three methods are  on average. ree cases, respectively; that of \textbf{B2 NN} converge to 6, 12.5, and 60; 6, 12.5, and 60 
% around 33\%, 66\%, 140\% 
% relative percentage rise.

% A. Convergence of violation. B. 1) the min; 2) 25\% quantile; 3) 50\%quantile; 4) 75\% quantile; 5) max. bullet item for a better illustration.
% \begin{itemize}
%     \item the min, it is the tail of the top, means  ;
%     \item 25\% quantile, it is the upper end of the boxes, it separates the lowest 25\% of the observations from the rest of the observations  ;
%     \item 50\% quantile;
%     \item 75\% quantile;
%     \item the maximum, it has .
%     \item outliners
%     \item consistent
% \end{itemize}
% the \textbf{Proposed NN} method has the maximum descent rate with the increase of training dataset size. Its median violation degrees converge to 4.5, 7.5, and 25 in the three cases, respectively. The \textbf{B2 NN} method's speed is middle, and its results converge to 6, 12.5, and 60 in the median, around 33\%, 66\%, 140\% 
% relative percentage rise. Moreover, for the third quartile (or 75th percentile),  the values between the two methods go from 5 to 15, 7 to 11.5, and 30 to 71, respectively. These numerical results indicate that worth-learning data generation improves the NN's generalization to ``unseen" data. The worst one is the \textbf{B1 NN} method. For the median and the third quartile, the violation degrees exceed that of the \textbf{Proposed NN} method by 56\%, 100\%, 220\%, and 250\%, 161\%, 467\%, respectively.

The comparison of optimality losses is similar to that of violation degrees, as illustrated in \cref{30bus-o:a,57bus-o:b,118bus-o:b}. The \textbf{proposed NN} method has the best results in the three cases, and the final median values of optimality losses are 0.6\%, 0.5\%, and 0.3\% in the three different cases, respectively. The optimality losses of \textbf{B2 NN} and \textbf{B1 NN} increase by 150\%, 66\%, and 360\% and 142\%, 167\%, and 460\% compared to those of the \textbf{proposed NN} method in the three cases. 

In conclusion, the proposed physical-model-integrated NN OPF solver with worth-learning data generation can improve the generalization of NN models compared to the conventional NN solvers. Specifically, the proposed method introduces an over 50\% reduction of constraint violations and optimality losses in the results on average.

\subsection{Comparison with numerical solvers}

To further evaluate the capability of the proposed method, the next experiment focuses on the comparison with the results of the classical AC-OPF solver based on the prime-dual interior points algorithm and the classical DC-OPF solver with a linear approximation of the power flow equations. The classical AC-OPF solver produces the optimal solutions as the ground truth values, and the DC-OPF solver is a widely used approximation in the power industry. The test set is the same as that in \cref{reNopt}. The performance of the three methods is evaluated by the following metrics: 1) the average consumption time to solve an OPF problem; 2) the average constraint violation degree $\boldsymbol{Vio}_\text{phm}$, {which is calculated by \cref{equ:phm1,equ:phm2}} for the two numerical solvers; and 3) the average relative error of dispatch costs. These three metrics are denoted as $\text{Time (ms)}, \text{Vio.} (\text{MW}),$ and $\text{Opt.} (\%)$, respectively. 

The results are tabulated in \cref{tab2}. The bottom row of the table shows the average results over the three cases. As shown, the proposed method achieves high computational efficiency, which is at least three orders of magnitude faster than the DC-OPF solver and four orders of magnitude faster than the AC-OPF solver. Furthermore, the method also has much lower constraint violations and optimality losses compared with the DC OPF solver.  
The average $\text{Vio.} 
~(\text{MW})$ and $\text{Opt.}~(\%)$ of the proposed solver are only 10.882 and 0.462, which are 44\% and 18\% of those of the DC-OPF solver, respectively.
% \textcolor{red}{However, the violation and computation speed of the proposed method is smaller than that of the DC linearization method but still larger than that of original AC solvers with orders of magnitude.}

% These results suggest that the proposed method is up to the precision better than the DC linearization method, but still worse than original AC solvers with orders of magnitude.

% \squeezetable
\renewcommand\arraystretch{0.9}
\begin{table*}[] \vspace{-3mm}
\small
\centering
\caption {Performance comparison of numerical solvers and the proposed solver} 
% \begin{ruledtabular}
\begin{tabular}{llllllllll}
 \toprule
 \multirow{2}{2cm}{Test \\ cases} & \multicolumn{3}{c}{AC-OPF solver} &  \multicolumn{3}{c}{DC-OPF solver}  &  \multicolumn{3}{c}{Proposed NN solver}   \\ 
\cmidrule(r){2-4}  \cmidrule(r){5-7}  \cmidrule(r){8-10} 
&Time (ms) &
  Vio. ($\text{MW}$) &
  Opt. (\%) & Time (ms) &
  Vio. ($\text{MW}$) &
  Opt. (\%) & Time (ms) &
  Vio. ($\text{MW}$) &
  Opt. (\%) \\
 
 \midrule
 30-bus                 & 530.3  & 0 & 0 & 14.8 & 5.340  & 0.908 & 0.110  & 4.415  & 0.603 \\
57-bus                 & 991.6  & 0 & 0 & 36.2 & 15.611 & 1.758 & 0.113  & 7.226  & 0.499 \\
118-bus                & 1606.7 & 0 & 0 & 78.5 & 52.199 & 4.762 & 0.116  & 21.004 & 0.285 \\
 \midrule
{Avg.}    & 1024.9 & 0 & 0 & 129.5 & 24.383 & 2.476 & 0.113 & {10.882} & {0.462}                                  \\
 \bottomrule
\end{tabular}
\label{tab2}\vspace{-5mm}
\end{table*}

\subsection{Interpretation of worth-learning data generation}
This subsection interprets why the worth-learning data generated by the proposed method improve the representativeness of the training dataset. 
The proposed worth-learning data generation method is compared with the conventional simple random sampling method. Without loss of generality, the experiment is conducted on the 14-bus system. Beginning with an identical initial dataset, the conventional and proposed methods generate 100 samples in every step, and there are 8 steps for both. To visualize the representativeness, we draw the distribution of these high-dimensional training samples based on the t-distributed Stochastic Neighbor Embedding algorithm\cite{scikit-learn,belkina2019automated}, which is a statistical method for visualizing high-dimensional data by giving each data point a location in a two- or three-dimensional map. 

\begin{figure}[t] \vspace{-2mm}%设置浮动属性，不设的话，图片可能会出现在页面的顶部，而不是在文字后面。
  \centering %居中
\subfigure[Training dataset generated by simple random sampling] { \label{dataRandom}
\includegraphics[width=1\columnwidth]{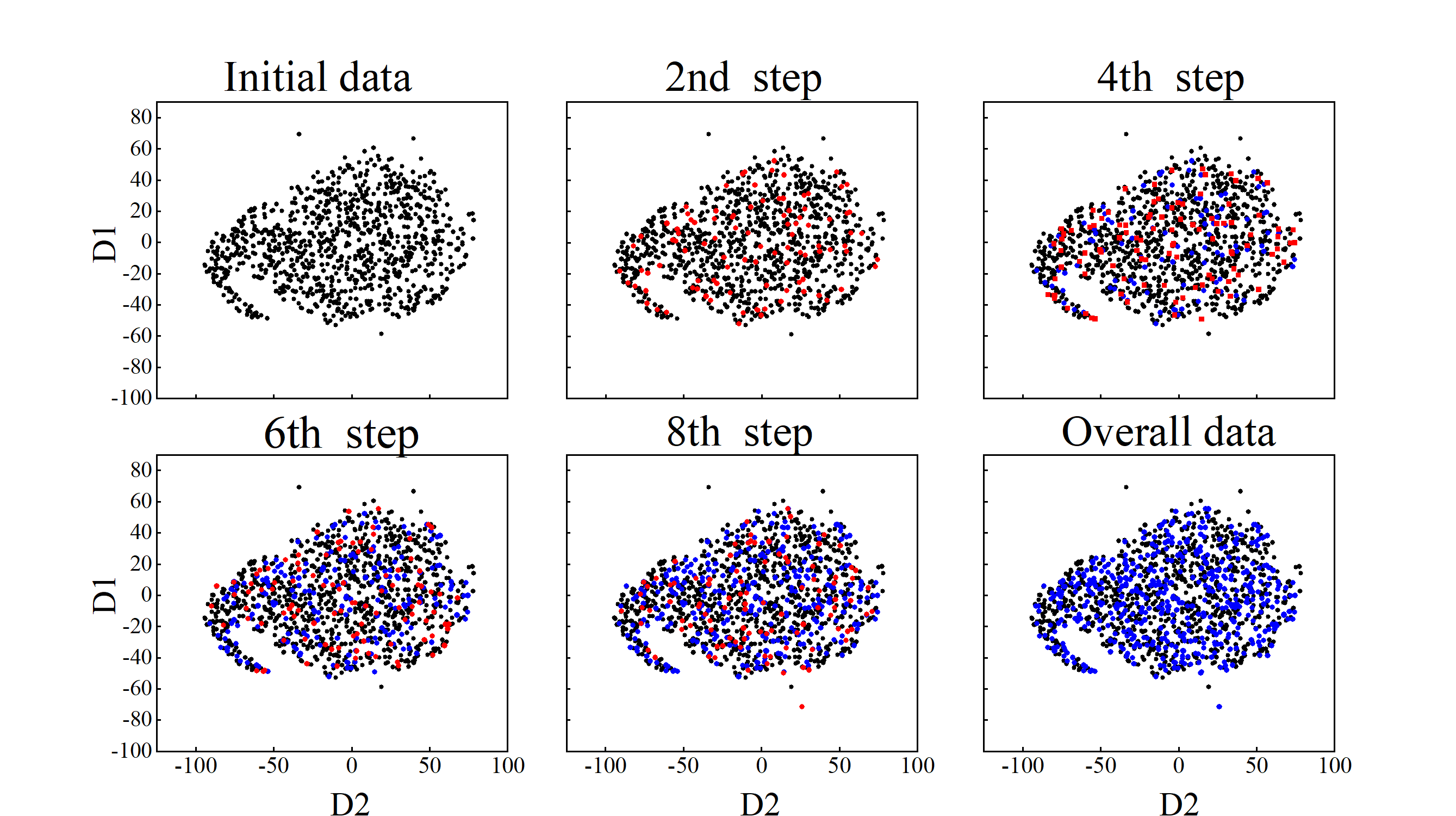}
}  
\subfigure[Training dataset generated by worth-learning data generation method] { \label{unfittedDat}
\includegraphics[width=0.94\columnwidth]{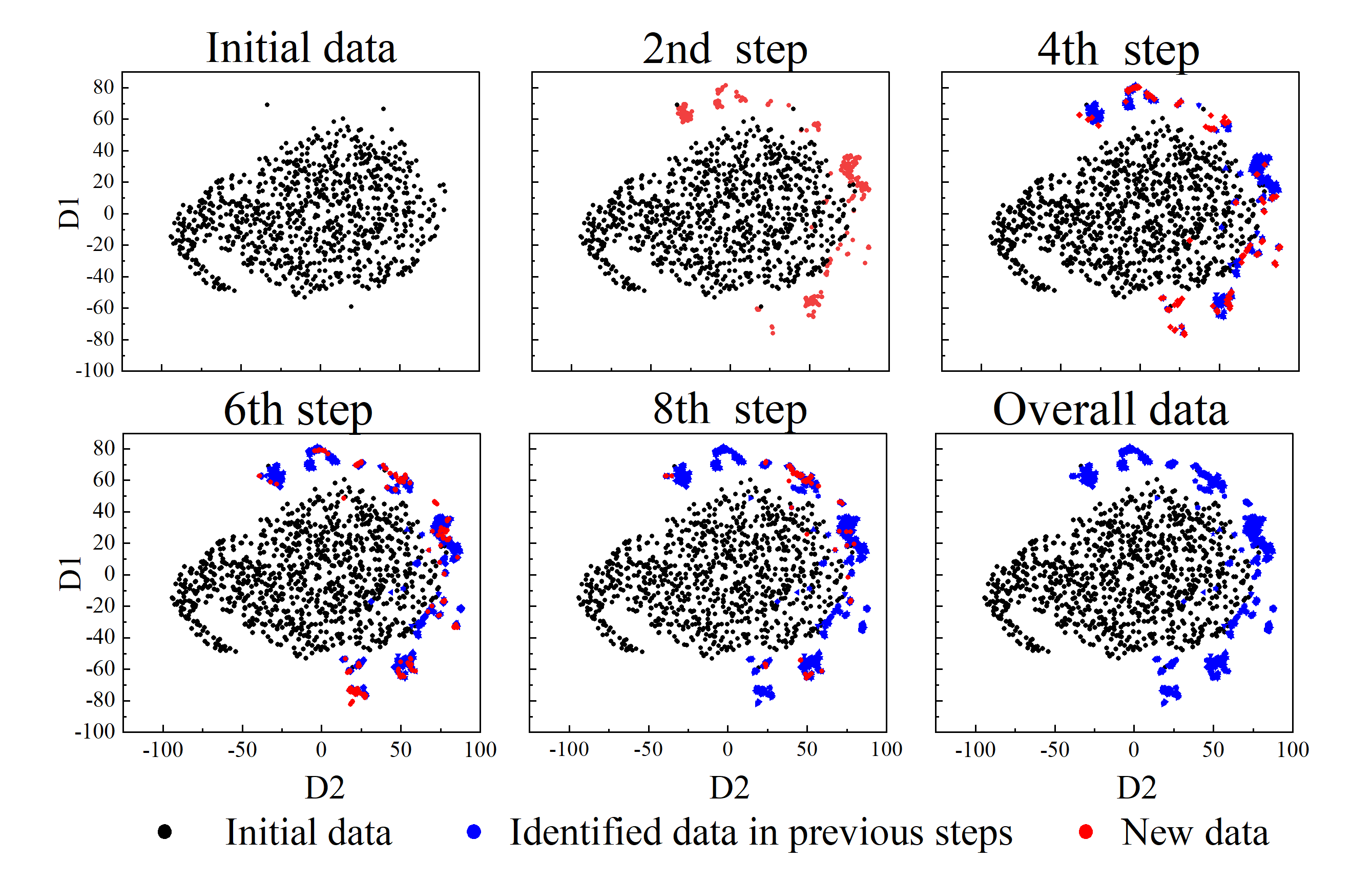}
}
\caption{Reduced-dimensional distributions of the training datasets generated by two different methods.}
\label{comparsionDataGeneration} \vspace{-5mm}
\end{figure}

The reduced-dimensional data distributions of the conventional and proposed methods are shown in Fig.\ref{comparsionDataGeneration}. In Fig. \ref{dataRandom}, the data are produced by the simple random sampling method, and their distribution is almost in a ``\faComment" region, which means the possibility of sampling in this region is high. Furthermore, the new data added in each step overlap with existing data or fill in the intervals. The new data overlapping with existing data are redundant in terms of NN training. The data filling in the intervals may be also redundant when the blanks are generalized well by the trained NN model. In contrast, as shown in Fig. \ref{unfittedDat}, the new data generated by the proposed method in each step hardly overlap with existing data and are usually outside the region covered by the initial data. {These new data increase the area covered by the training set so that the training set can have better representativeness of the input feasible region. This explains the effectiveness of the proposed worth-learning data generation method.} 
% \textcolor{red}{Compared to the conventional method, the worth-learning data generation usually produces data that are distinguishable from existing ones. These data increase the representativeness of the training set by expanding the area of the training data distribution.}

% \textcolor{red}{The generalization of NNs is negatively correlated with the distance from the training data. So we can know that the max violation backpropagation particularly finds the points that are badly generalized by NNs, which indicates these points are worth learning.}

% \vspace{-10mm}

% \subsection{conclusions}
\section{Conclusion}
\label{Section 6}
{This study proposes an AC-OPF solver based on a physical-model-integrated NN with worth-learning data generation to produce reliable solutions efficiently. To the best of our knowledge, this is the first study that has addressed the generalization problem of NN OPF solvers regarding the representativeness of training datasets. 
The physical-model-integrated NN is designed by integrating an MLP and an OPF-model module. This specific structure outputs not only the optimal decision variables of the OPF problem but also the constraint violation degree. Based on this NN, the worth-learning data generation method can identify feasible training samples that are not well generalized by the NN. Accordingly, by iteratively applying this method and including the newly identified worth-learning data samples in the training set, the representativeness of the training set can be significantly enhanced. }

{The theoretical analysis shows that the method brings little computational burden into the training process and can make the models generalize over the feasible region.}
% Many engineering techniques for transforming the OPF models into the \textit{physical model module} and \textit{input feasible set module} are displayed, which is necessary for readers to understand and reproduce the process.}
Experimental results show that the proposed method leads to over a 50\% reduction of both constraint violations and optimality loss compared to conventional NN solvers. Furthermore, the computation speed of the proposed method is three orders of magnitude faster than that of the DC-OPF solver.

\appendices

\section{Computational Efficiency of Worth-Learning Data Generation}
\label{appendix:efficiencyofMVB}

To analyze the computational complexity of the proposed NN model with worth-learning data generation, we adopt a widely used measure---the number of floating-point operations (FLOPs) during the NN model's forward-backward propagation. The total FLOPs of one single layer of a fully-connected NN model can be calculated as follows:
\begin{subequations}
\label{FLOP_fun}
\begin{gather}
    \textbf{Forward}: \quad \text{FLOPs}=(2I-1)\times O, \\
    \textbf{Backward}: \quad \text{FLOPs}=(2I-1)\times O,
\end{gather}
\end{subequations}
where $I$ is the dimension of the layer's input, and $O$ is the dimension of its output.

To approximate an OPF mapping based on a 57-bus system, the proposed NN model uses the following structure: $84\times 1000\times2560 \times2560\times 5120\times 2000 \times114 $. According to \cref{FLOP_fun}, the total FLOPs of the NN per forward-backward process is around $1\times10^8 $. The GPU used in the experiment is the Quadro P6000, and its performance is 12.2 TFLOP/s ($1 \ \text{TFLOP/s}=10^{12} \text{ FLOP/s}$).  Using the GPU,  we can perform the forward-backward process $1.22 \times 10^{5}$ times per second. 

For the worth-learning data generation in \cref{alg:the_alg}, the forward process is to calculate $\boldsymbol{Vio}_{\text{ifs}}$ and $\boldsymbol{Vio}_{\text{phm}}$, and the backward process is to update $\boldsymbol{S'}^\text{G}_\text{ifs}$ and $ \boldsymbol{V'}_\text{ifs}$ by the gradients. We concatenate $\boldsymbol{S'}^\text{G}_\text{ifs}$ and $ \boldsymbol{V'}_\text{ifs}$ as a vector $\textbf{x}$, and we suppose the range of each item in $\textbf{x}$ is $[0, 10]$, and $\textbf{x}$ changes $10^{-3}$ in each update step. Varying from 0 to 10, it costs $10^4$ times the forward-backward processes. In other words, the algorithm can at least update $1.22 \times 10^{5} /
10^4\approx12 $ samples in 1 s, so finding one sample costs no longer than 0.08 s.

In practice, there is a slight error between the actual speed in experiments and the theoretical analysis. According to the numerical experiments in \cref{section:effi}, an average of 533 samples are found in 30 s. The average consumption time for identifying one sample is 0.056 s. 

From the analysis presented above, we can conclude that the proposed worth-learning data generation method brings little computational burden into the training process.

\section{Convergence of Worth-Learning Data Generation}
\label{appendix:convergenceMVB}
This section verifies that the proposed NN with worth-learning data generation can generalize to the whole feasible set. NN models are continuous functions because both linear layers and activation functions are continuous. We define a critical violation value $\epsilon$ that divides the input space into two types: the covered region (the $\boldsymbol{Vio}_\text{phm}$ values of all of the points are less or equal to $\epsilon$) and the uncovered region (the $\boldsymbol{Vio}_\text{phm}$ values of all of the points  are greater than $\epsilon$). The boundaries of the two regions consist of the points whose $\boldsymbol{Vio}_\text{phm}$ values are approximately equal to $\epsilon$. Using these points as initial points, we can identify points with the local maximum in the uncovered region by max violation backpropagation.

\begin{figure}[tb]
    \centering
    \includegraphics[width=2.8in]{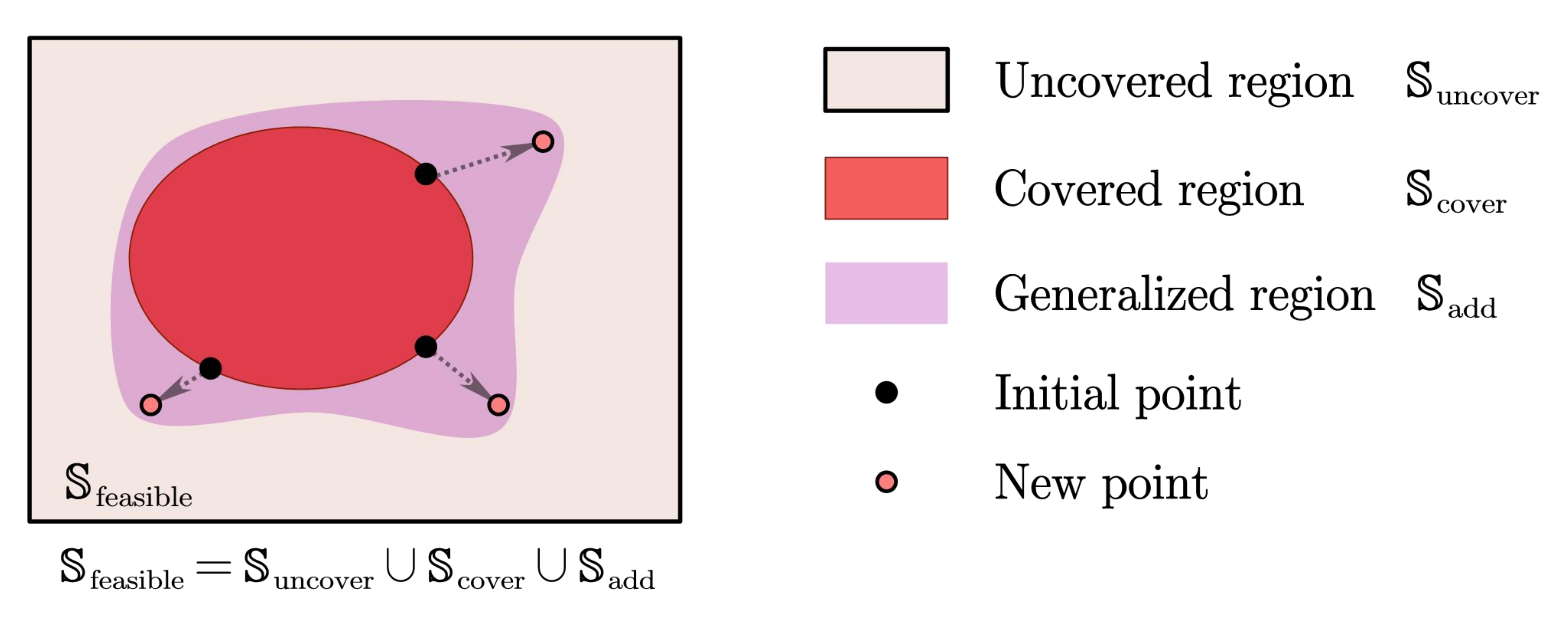}
    \caption{Illustration of the covered region $\mathbb{S}_\text{cover}$ expanding its area by the generalized region $\mathbb{S}_\text{add}$.}
    \label{convengence} \vspace{-5mm}
\end{figure}

{Next, these new points $\{\boldsymbol{x}_\text{1}\} $ (the red points) are added to the training set. After training,  the neighborhood of these new points $\{\boldsymbol{x}_\text{1}\}$ would be covered. Due to the generalization of NNs, most points in the area $\mathbb{S}_\text{add}= \{\boldsymbol{x}|a\times \boldsymbol{x}_\text{0}^\text{ini}+(1-a)\times \boldsymbol{x}_\text{1}, 0\le a \le 1  \}$ would also be covered, where $\{\boldsymbol{x}_\text{0}^\text{ini}\}$ are the initial points on the boundaries (the black points), as shown in \cref{convengence}.}

Therefore, the area $\mathbb{S}_\text{add}$ is subtracted from the uncovered region. Through iterations, the uncovered region is emptied, and the number of added samples converges to zero.

In practice, we choose the training set instead of the boundary points as initial points for convenience. Although some samples in the training set are not at boundaries, they are eliminated 
 by the filter function, as shown in \cref{alg:the_alg}. Therefore, the replacement of the boundary points has no impact on the results.
 % Other train samples are also used. Therefore, some new points' violations are not greater than $\epsilon $. For this situation, a filter function is employed on $x_0^\text{ini}$ and $x_{1}$ before the next iteration, as shown in line 13 of \cref{alg:the_alg}, then the initial points in the next iteration consist of filtered previous and added samples.

\bibliographystyle{IEEEtran}
\bibliography{IEEEabrv, refers}

\end{document}